\pdfoutput=1

\documentclass[11pt]{article}

\usepackage[preprint]{acl}
\usepackage{tikz}
\usetikzlibrary{shapes.geometric, positioning, arrows.meta, decorations.pathmorphing}

\usepackage{times}
\usepackage{latexsym}
\usepackage{bbm}
\usepackage[T1]{fontenc}
\usepackage[utf8]{inputenc}
\usepackage{amsmath}
\usepackage{amsfonts}
\usepackage[inkscapelatex=false]{svg}
\usepackage{microtype}
\usepackage{booktabs}
\usepackage{algorithm}
\usepackage{algpseudocode}
\usepackage{inconsolata}
\usepackage{graphicx}
\usepackage{multirow}
\usepackage{enumitem}
\usepackage{balance}

\title{Machine-generated text detection prevents language model collapse}

\author{George Drayson{\normalfont,} Emine Yilmaz {\,\normalfont and\,} Vasileios Lampos\\
  Centre for Artificial Intelligence\\
  Department of Computer Science\\
  University College London, UK\\
  \{\texttt{george.drayson.23},\,\texttt{emine.yilmaz},\,\texttt{v.lampos}\}\texttt{@ucl.ac.uk}}
  
\begin{document}

\maketitle

\begin{abstract}
As Large Language Models (LLMs) become increasingly prevalent, their generated outputs are proliferating across the web, risking a future where machine-generated content dilutes human-authored text. Since online data is the primary resource for LLM pre-training, subsequent models could be trained on an unknown portion of synthetic samples. This could lead to model collapse, a degenerative process whereby LLMs reinforce their own errors, reduce output diversity, and ultimately yield declining performance. In this study, we investigate the impact of decoding strategy on model collapse, analysing the text characteristics at each model generation, the similarity to human references, and the resulting model performance. Using the decoding strategies that lead to the most significant degradation, we evaluate model collapse in a more realistic scenario where the origin of the data (human or synthetic) is unknown. We train a machine-generated text detector and propose an importance resampling approach to prevent model collapse by up-sampling likely human content in the training data. Our method is validated on four LLMs from two model families (\texttt{GPT-2} and \texttt{SmolLM2}), across a range of model sizes ($124$M to $1.7$B). We demonstrate that it not only prevents model collapse but also improves performance compared to training on purely human data, underscoring the benefit of synthetic samples and the importance of data curation.\footnote{\textbf{Accepted by EMNLP 2025} (main).}

\textbf{Source code:} \href{https://github.com/GeorgeDrayson/model_collapse}{ \nolinkurl{github.com/GeorgeDrayson/model\_collapse}}

\end{abstract}

\section{Introduction}
Large Language Models (LLMs) can generate high-quality, fluent language across a wide range of applications. A key factor that drives their capabilities is the vast amount of data used to train them, which is predominantly based on text published on the web \cite{wenzeketal-2020-ccnet}. The extensive adoption of LLMs will inevitably result in an ever-increasing amount of synthetic data that will co-exist alongside or even dominate human-generated text~\cite{dohmatob2024tale}, especially within online ecosystems such as social media, news websites, and digital encyclopedias. Hence, there are legitimate concerns as to the effect this might have on future generations of language models trained on a mixed set of human and synthetic corpora.

While synthetic data has proven beneficial in controlled scenarios, such as instruction tuning~\cite{wang2023self}, distillation~\cite{hsieh-etal-2023-distilling-custom}, self-improvement~\cite{wu2025metarewarding}, and reinforcement learning~\cite{zhao2025absolute}, these settings typically involve careful curation and limited reuse. In contrast, our focus is on the long-term effects of uncontrolled accumulation of synthetic content. Several works have attempted to simulate this scenario by recursively training language models on their own generated outputs~\cite{shumailov2023curse, briesch_large_2024, guo-etal2024-curious}. The outcome of this recursive training is referred to as ``\emph{model collapse}''~\cite{shumailov2023curse}, a degenerative process caused by training on synthetic data from previous generations, leading to compounded errors and the convergence to a low variance output distribution. This has been shown to cause performance degradation~\cite{alemohammad_self-consuming_2023} and a drastic loss in diversity~\cite{briesch_large_2024, guo-etal2024-curious, alemohammad_self-consuming_2023}. However, an unexplored factor in this recursive training process is the decoding strategy used to generate the synthetic data. Decoding strategies alter the distribution of model outputs, which could, in turn, affect how errors accumulate during recursive training.

This work investigates the impact of decoding strategies on model collapse and the characteristics of the data that could be causing this behaviour. Subsequently, we explore the scenario where the training data is mixed (human and synthetic) in an unknown proportion, akin to training on web-crawled data. We propose a method for preventing model collapse by a guided resampling of the training data using the confidence estimates from a machine-generated text detector. Our method is motivated by prior work~\cite{bertrand2024on,alemohammad_self-consuming_2023}, which highlighted that when the proportion of human data in the training set is sufficient, model collapse can be prevented.

Our contributions can be summarised as follows:
\begin{enumerate}[label=(\alph*), leftmargin=*, topsep=0.5pt, itemsep=-4pt]
    \item we evaluate model collapse from three perspectives: task performance, model generation quality, and semantic similarity to human text,
    \item we show that model collapse is significantly affected by the choice of decoding strategy, demonstrating large discrepancies in performance and data quality,
    \item we train a machine-generated text detector to provide calibrated confidence estimates for the origin of the training samples,
    \item we propose a method that uses the detector's outputs to prevent model collapse through a guided resampling of the training data, and
    \item we substantiate our hypotheses by conducting experiments on four LLMs from two model families across a range of decoding strategies.
\end{enumerate}

\section{Prior work on model collapse}
\label{sec:model_collapse}
Model collapse is a degenerative process in which models recursively trained on generational data exhibit a drop in performance compared to a model trained on the original human distribution~\cite{shumailov2023curse}. In the early stages of model collapse, information is lost at the tails of the distribution. Eventually the output distribution converges to a point estimate with very little variance, resulting in a model that cannot be restored back to the original generation trained on human data. This effect can also be viewed as a change to neural scaling laws, in which there reaches a point where training on additional synthetic samples does not improve model performance, and thus learning plateaus~\cite{dohmatob2024tale}.

It has been argued that the two causes for this behaviour are finite sampling error leading to information being lost at the tails of the distribution, and functional approximation error introducing non-zero likelihoods outside of the support of the original distribution~\cite{shumailov2023curse}. Additionally, \citet{dohmatob2024tale} theorised that the choice of generation algorithm is another contributing factor to model collapse. However, this has not been empirically evaluated in the case of LLMs, where decoding strategies that modify and sample from the model's output distribution could also have a significant impact on how errors are propagated across generations. Currently, model collapse in LLMs has been studied with a fixed decoding strategy, and model degradation has been mostly assessed using task performance metrics such as perplexity~\cite{shumailov2024ai} and test loss~\cite{gerstgrasser_is_2024}. Interestingly,~\citet{guo-etal2024-curious} also evaluate the diversity of the generated text. In our study, we have chosen to study model collapse across three perspectives: the quality of the generated text (including diversity and readability), its similarity to human text, and the downstream task performance.

Recent studies have explored methods for mitigating model collapse, such as using synthetic samples as negative guidance in the image domain~\cite{alemohammad_self_improving_2024}, pruning samples based on high perplexity~\cite{feng_beyond_2024}, token-level editing \cite{zhu2024synthesize} or filtering low-quality samples~\cite{zhang_regurgitative_2024}. \citet{bertrand2024on}~and~\citet{alemohammad_self-consuming_2023} show that when a high enough proportion of human data is added at each iteration, model collapse in diffusion models can be avoided. In the computational linguistics domain,~\citet{gerstgrasser_is_2024} showed that by accumulating all cross-generational data and combining it with the original human data, model collapse can be mitigated. However, in these works, the models are trained on either entirely synthetic data or the true labels of the samples are known a priori. In our work, we investigate how to prevent model collapse in a more realistic setting where the training data is mixed and the origin (human or synthetic) of the samples is unknown.

\section{Background}
\label{subsec:language_model}
In this work, we study open-ended text generation, in which a token sequence, 
$\mathbf{x}\!=\!\{x_1,\dots,x_m\}$, is provided as context to a language model and the task is to generate a continuation, $\mathbf{\hat{x}}\!=\!\{\hat{x}_{1},\dots,\hat{x}_{c}\}$, from the model's probability distribution, $p_{\theta}\!\left(\hat{\mathbf{x}}\right)$, where $\theta$ denotes the model's parameters:
\begin{equation}
    p_{\theta}\!\left(\hat{\mathbf{x}}\right) = \prod_{i=1}^{c} p_{\theta}\!\left(\hat{x}_i\mid \{\mathbf{x},\hat{\mathbf{x}}_{<i}\}\right).
    \label{eq:language_modeling}
\end{equation}
Tokens are selected from the probability distribution at each step by following a decoding strategy, resulting in a text sample $\{\mathbf{x}, \hat{\mathbf{x}}\}$. There are two main categories of decoding strategies, deterministic and stochastic. The former is designed to maximise the joint probability of the generated sequence, e.g. by selecting the most probable token at each step (\textbf{greedy decoding}) or keeping track of multiple candidate text sequences and selecting the most probable (\textbf{beam search}).
Stochastic methods, on the other hand, sample from the model's output distribution, resulting in less repetitive and more human-like text~\cite{Holtzman2020The}. The simplest stochastic method, \textbf{pure sampling}, samples directly from the distribution $p_{\theta}$. \textbf{Top-$k$ decoding}~\cite{fanetal-2018-hierarchical}, samples from the $k$ most probable tokens to avoid text generation from the tail of $p_{\theta}$. A more nuanced approach, \textbf{nucleus sampling}~\cite{Holtzman2020The}, dynamically truncates the vocabulary to the highest probability tokens by thresholding the cumulative probability mass with a parameter $\eta\!\in\![0, 1]$. In addition, the probability mass can be skewed towards high-probability outcomes by deploying \textbf{temperature}, controlled by $\tau\!\in\![0, 1]$~\cite{ackley_learning_1985}.

\section{Methods}
\label{sec:methods}
In this section, we provide an overview of the methods and metrics used in our experiments, including the details of the machine-generated text detector.

\subsection{Recursive LLM training}
\label{subsec:recursive_training}
Similarly to~\citet{shumailov2024ai}~and~\citet{dohmatob2024tale}, we simulate model collapse by fine-tuning a language model recursively on its own generated output (entirely or partially, depending on our underlying hypothesis) for a fixed number of generations. This process is described in Algorithm~\ref{alg:recursive_training}. Recursive training commences by fine-tuning a pre-trained language model, $p_{\theta}$, using a dataset consisting of $n$ human-generated samples, $\mathcal{D}_{\text{H}}\!=\!\{\mathbf{x}_j\}_{j=1}^{n}$. This results in a model $p^0$, where `$0$' denotes the stage of the entire process (generation).\footnote{For enhanced notational clarity, we choose to drop parameter $\theta$ for the recursively produced LLMs. However, we clarify that $\theta$ is updated in each generation.} We then use a set of $n$ context sequences, $\mathcal{X}\!=\!\{\mathbf{x}_1, \dots, \mathbf{x}_n\}$ (one for each sample in $\mathcal{D}_{\text{H}}$), to generate a set of continuation sequences, $\mathcal{\hat{X}}\!=\!\{\mathbf{\hat{x}}_1, \dots, \mathbf{\hat{x}}_n\}$, where $\hat{\mathbf{x}}_j\!\sim\!p^0$ (see also section~\ref{subsec:language_model}).
The human-generated context together with the LLM-generated continuation sequences form a new synthetic dataset, $\mathcal{D}_{\text{S}}^1$ (here `$1$' is used to denote that this dataset will be used to fine-tune a language model in the next generation).

Subsequently, successive rounds of recursive training are carried out. In each generation $i$, the original language model, $p_{\theta}$, is fine-tuned using the synthetic dataset $\mathcal{D}_{\text{S}}^i$ to obtain $p^i$. Thereafter, $p^i$ is prompted with context sequences $\mathcal{X}$ to generate a new synthetic dataset $\mathcal{D}_{\text{S}}^{i\!+\!1}$ that will be used to fine-tune $p_{\theta}$ in generation $i\!+\!1$.

\begin{algorithm}[t]
\caption{Recursive LLM training}
\label{alg:recursive_training}
\begin{algorithmic}[1]
\State \textbf{Input:} Human text samples $\mathcal{D}_{\text{H}}\!=\!\{\mathbf{x}_j\}_{j=1}^{n}$, pre-trained language model $p_{\theta}$ \hfill
\State Obtain $p^0$ by fine-tuning $p_{\theta}$ using $\mathcal{D}_{\text{H}}$ \hfill
\For{$i\!=\!1, \dots, G$}
    \State $\mathcal{D}_{\text{S}}^{i}\!=\!\{\mathbf{x}_j, \hat{\mathbf{x}}_j\}_{j=1}^{n} , \text{ where } \hat{\mathbf{x}}_j \sim p^{i-1}$ \hfill
    \State Obtain $p^i$ by fine-tuning $p_{\theta}$ using $\mathcal{D}_{\text{S}}^{i}$ \hfill
\EndFor
\State \textbf{Outputs:} $p^i \, (i\!\ge\!0)$, $\mathcal{D}_{\text{S}}^{i} \, (i\!\ge\!1)$
\end{algorithmic}
\end{algorithm}

\subsection{Metrics for LLM performance}
\label{subsec:LLM_performance}
We evaluate model collapse by fine-tuning and testing models on the open-ended text generation task, emulating the setup proposed by~\citet{shumailov2024ai}. We assess language model performance in terms of perplexity and accuracy. \textbf{Perplexity} measures how well the model predicts unseen text, with lower values indicating better performance. \textbf{Accuracy}, in this context, reflects the proportion of correctly predicted tokens, providing a direct measure of the model's effectiveness in generating accurate language outputs.

\begin{figure*}[t]
    \centering
    \includegraphics[width=0.86\linewidth]{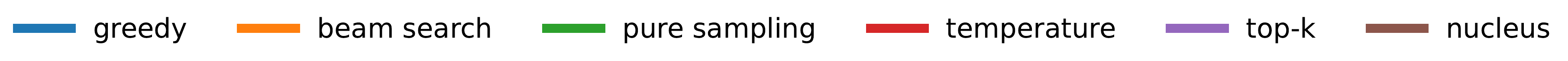}
    
    \begin{tabular}{c@{\hspace{0.2cm}}c@{\hspace{0.5cm}}c@{\hspace{0.2cm}}c@{}}
        \includegraphics[width=0.23\linewidth]{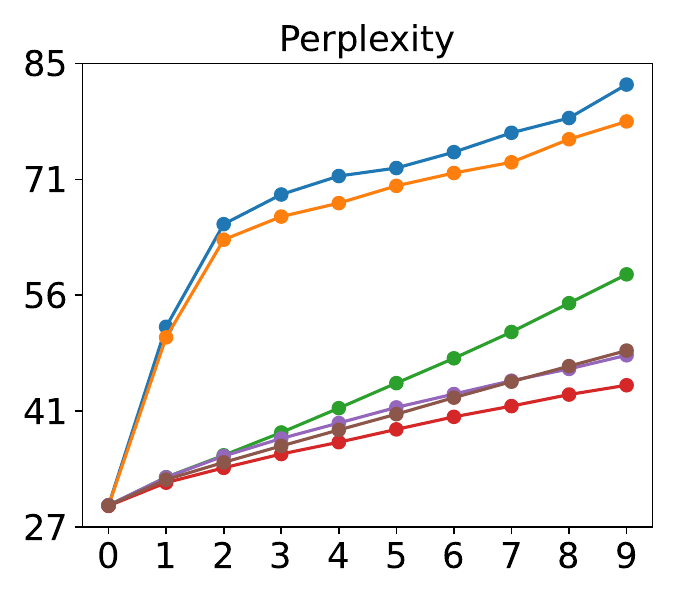} &
        \includegraphics[width=0.23\linewidth]{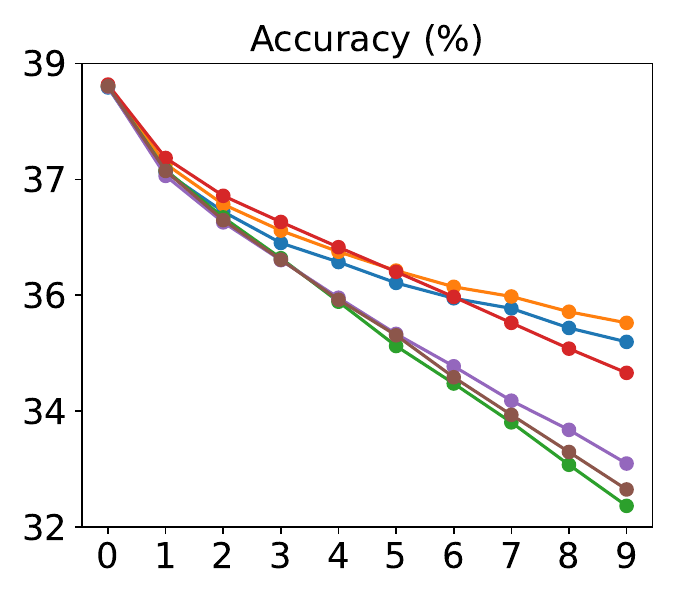} &
        \includegraphics[width=0.23\linewidth]{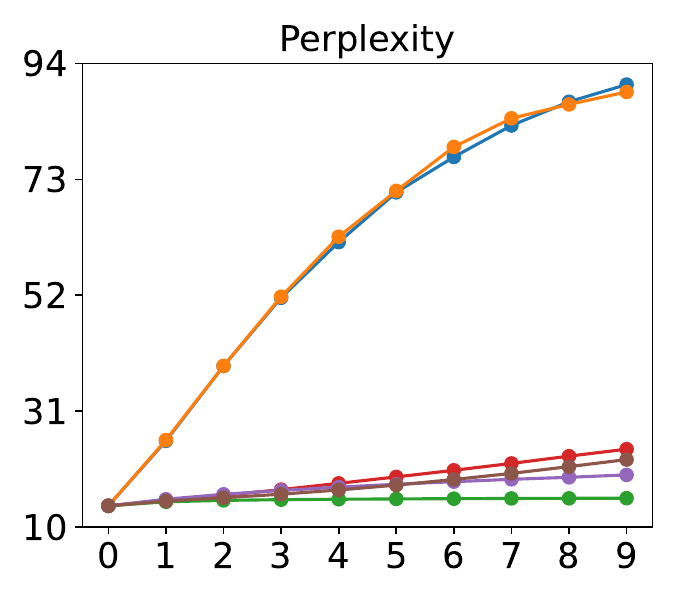} &
        \includegraphics[width=0.23\linewidth]{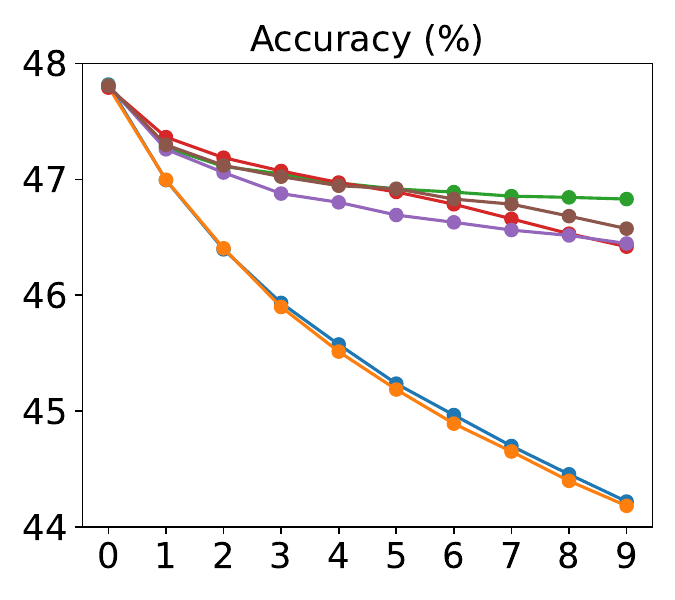}\\
        \multicolumn{2}{c}{\scriptsize\texttt{GPT-2}} & \multicolumn{2}{c}{\scriptsize\texttt{SmolLM2}}
    \end{tabular}
    \caption{Perplexity and accuracy over generations $0$ to $9$ of fully synthetic recursive training.}
    \label{fig:fully_synthetic}
\end{figure*}

\subsection{Metrics for LLM text generation quality}
\label{subsec:qualitative_metrics}
We complement performance metrics with more qualitative ones drawn on the generated text outputs and their similarity to human references, to obtain a holistic understanding of LLM collapse.\\[2pt]
\textbf{Diversity} (\texttt{D}) takes into account the sequence-level repetition at different $n$-gram levels of a document. Higher scores are reflective of more lexically diverse text. We use the following formulation:
\begin{equation}
\texttt{D}\!\left(\hat{\mathbf{x}}\right)=\prod_{n=2}^{4} \left( 1 - \frac{\lvert \text{unique } n\text{-grams}(\hat{\mathbf{x}}) \rvert}{\lvert \text{total } n\text{-grams}(\hat{\mathbf{x}}) \rvert} \right) \, .
\end{equation}
\textbf{Self-BLEU}~\cite{zhu2018texygen} evaluates the BLEU score \cite{papineni2002bleu} of each document compared to all other documents in the generation set, providing a metric for how repetitive the model is for different outputs. We use a random sample of $1{,}000$ documents and evaluate Self-BLEU up to an $n$-gram size of $4$. A lower score indicates higher text diversity.\\[2pt]
\textbf{MAUVE}~\cite{pillutla2021mauve} measures the distribution similarity between the original human text and the generated text. It is computed using the Kullback–Leibler (KL) divergence between the two text distributions in the embedding space of an LLM. To perform this, we use a random sample of $1{,}000$ documents of human and machine-generated text. A higher score indicates that the model generated text with a stronger resemblance to human writing.\\[2pt]
\textbf{Readability} is evaluated using the Flesch-Kincaid Reading-Ease score \cite{flesch1948new}, which estimates how difficult it is to understand a passage based on the number of words, sentences, and syllables. We implement the metric using the \texttt{textstat} package.\footnote{\texttt{textstat} Python package, \href{https://textstat.org/}{\nolinkurl{textstat.org}}} Lower scores indicate more complex text, typically characterised by longer sentences and higher lexical density.

\subsection{Machine-generated text detection}
\label{subsec:detector}
Machine-generated text detection approaches can be divided into metric-based~\cite{mitchell2023detectgpt,hans2024spotting} and neural-based~\cite{hu2023radar,bhattacharjee2023conda}. The former use statistical features, often extracted from surrogate LLMs, to detect machine-generated text, whereas the latter are based on machine learning, such as fine-tuning a small pre-trained language model with a binary classification head. Here we deploy a neural classifier due to reported state-of-the-art (SOTA) performance on relevant machine-generated text detection benchmarks~\cite{wang2024semeval, mage} and the ability to approximate confidence estimates from the model's outputs.

Our detector is based on an encoder-only transformer model with a sequence classification head that maps the \texttt{CLS} token representation to logits, $\mathbf{z}_i$, which are converted to pseudo-probabilities using a sigmoid function, $\sigma$. As LLM training is considerably resource-intensive, any data filtering or sampling methods must be able to efficiently process large quantities of data with minimal computational overhead~\cite{wenzeketal-2020-ccnet}. With this in consideration, we evaluate the base variants of $3$ pre-trained language models with under $200$ million parameters: \texttt{RoBERTa}~\cite{liu_roberta_2019} and \texttt{DeBERTav3}~\cite{he2023debertav} due to their SOTA performance in machine-generated text detection~\cite{mage, wangetal-2024-m4gt}, and \texttt{ModernBERT}~\cite{warner_smarter_2024}, a more recent variant that has achieved superior performance on a range of natural language processing benchmarks. The added advantages of \texttt{ModernBERT} is the large context window ($8{,}192$ tokens), superior computational speed, and memory efficiency \cite{warner_smarter_2024}. See Appendix~\ref{sec:ai_text_detection} for more details.

Despite their overall strong detection performance, as with all modern neural networks, the confidence estimates are poorly calibrated~\cite{guo2017calibration}, i.e. they are not representative of the true likelihood. To mitigate overconfidence in the predictions, we apply label smoothing. Additionally, we use temperature scaling to further calibrate the model's predictions. Given the logit vector $\mathbf{z}_i$, the new confidence prediction is $\sigma\! \left(\mathbf{z}_i / T\right)$, where $T$ is a learnable temperature parameter.

\begin{table*}[t]
\centering
\small
\setlength{\tabcolsep}{2pt}
\resizebox{0.9\linewidth}{!}{
\begin{tabular}{llcccccccccccc}
\toprule
\multirow{2}{*}{\bf Model} & \multirow{2}{*}{\bf Decoding} & \multicolumn{2}{c}{\bf Perplexity $\downarrow$} & \multicolumn{2}{c}{\bf Accuracy $\uparrow$} & \multicolumn{2}{c}{\bf Diversity $\uparrow$} & \multicolumn{2}{c}{\bf Self-BLEU $\downarrow$} & \multicolumn{2}{c}{\bf MAUVE $\uparrow$} & \multicolumn{2}{c}{\bf Readability $\uparrow$} \\
& & Gen $0$ & Gen $9$ & Gen $0$ & Gen $9$ & Gen $0$ & Gen $9$ & Gen $0$ & Gen $9$ & Gen $0$ & Gen $9$ & Gen $0$ & Gen $9$ \\
\midrule
\multirow{6}{*}{\texttt{GPT-2}} 
& greedy & $29.29$ & $82.74$ & $38.72$ & $34.93$ & $0.96$ & $0.70$ & $61.02$ & $67.13$ & $0.99$ & $1.00$ & $60.47$ & $8.25$ \\
& beam search & $29.25$ & $78.06$ & $38.75$ & $35.21$ & $16.78$ & $10.86$ & $61.54$ & $67.60$ & $0.91$ & $1.29$ & $60.97$ & $17.57$ \\
& pure sampling & $29.29$ & $58.64$ & $38.74$ & $32.49$ & $94.88$ & $99.82$ & $24.12$ & $6.76$ & $90.16$ & $7.18$ & $40.62$ & $-10.14$ \\
& temperature & $29.23$ & $44.55$ & $38.77$ & $34.47$ & $87.76$ & $25.10$ & $33.45$ & $54.56$ & $94.15$ & $22.69$ & $46.80$ & $\mathbf{36.79}$ \\
& top-$k$ & $29.31$ & $48.36$ & $38.73$ & $33.12$ & $84.57$ & $70.20$ & $38.81$ & $\mathbf{42.14}$ & $95.21$ & $\mathbf{70.01}$ & $51.19$ & $34.32$ \\
& nucleus & $29.28$ & $48.96$ & $38.74$ & $32.73$ & $92.26$ & $\mathbf{86.73}$ & $28.24$ & $26.86$ & $90.96$ & $57.41$ & $43.96$ & $21.31$ \\
\midrule
\multirow{6}{*}{\texttt{SmolLM2}}
& greedy & $13.96$ & $85.69$ & $47.58$ & $43.76$ & $6.68$ & $2.22$ & $57.54$ & $50.12$ & $3.23$ & $0.99$ & $62.98$ & $47.04$ \\
& beam search & $13.96$ & $84.75$ & $47.59$ & $43.77$ & $6.64$ & $2.16$ & $57.30$ & $50.38$ & $3.06$ & $0.89$ & $62.69$ & $45.87$ \\
& pure sampling & $13.96$ & $15.39$ & $47.58$ & $46.59$ & $90.74$ & $\mathbf{90.72}$ & $47.23$ & $\mathbf{45.66}$ & $86.00$ & $\mathbf{82.80}$ & $47.55$ & $\mathbf{47.59}$ \\
& temperature & $13.96$ & $24.88$ & $47.55$ & $46.05$ & $82.92$ & $24.81$ & $51.15$ & $57.55$ & $89.94$ & $17.60$ & $52.83$ & $64.09$ \\
& top-$k$ & $13.96$ & $19.86$ & $47.59$ & $46.02$ & $82.72$ & $56.77$ & $52.62$ & $57.91$ & $85.97$ & $59.84$ & $55.52$ & $63.18$ \\
& nucleus & $13.96$ & $22.81$ & $47.58$ & $46.23$ & $87.33$ & $44.27$ & $49.62$ & $58.39$ & $92.39$ & $53.00$ & $51.20$ & $64.17$ \\
\midrule
Human & & \multicolumn{2}{c}{—} & \multicolumn{2}{c}{—} & \multicolumn{2}{c}{$88.79$} & \multicolumn{2}{c}{$42.89$} & \multicolumn{2}{c}{$100$} & \multicolumn{2}{c}{$50.34$} \\
\bottomrule
\end{tabular}
}
\caption{Impact of decoding strategies on the model performance and text generation quality (comparison between generations $0$ and $9$) in the fully synthetic recursive training setting. \textbf{Bold font} indicates the closest score to the human reference for generation $9$ ($\uparrow$ / $\downarrow$: higher / lower is better).}
\label{tab:data_quality_fully_synthetic}
\end{table*}

\section{The impact of decoding strategies on model collapse}
\label{sec:exp_decoding}
We carry out recursive training as described in section \ref{subsec:recursive_training} on the open-ended text generation task by fine-tuning \texttt{GPT-2} $124$M ~\cite{radford2019language} and
\texttt{SmolLM2} $360$M~\cite{smollm2} on the WikiText-2 dataset~\cite{merity2016pointer}. The Wikipedia articles are segmented into non-overlapping chunks of $512$ tokens, where the first $256$ are used as the context ($\mathbf{x}$), and the remaining $256$ as the continuation ($\hat{\mathbf{x}}$). We conduct full fine-tuning for $1$ epoch and, to avoid information leakage between generation and training, define cross-entropy loss only on the generated sequence of each sample, $\mathcal{\hat{X}}\!=\!\{\mathbf{\hat{x}}_1, \dots, \mathbf{\hat{x}}_n\}$~\cite{dohmatob2024tale}. Additional details can be found in Appendix~\ref{sec:model_collapse_training}. We evaluate a range of decoding strategies to assess the effect on model collapse: greedy decoding ($\tau=0$), $5$-way beam search, pure sampling ($\tau=1$), temperature ($\tau=0.9$), top-$k$ (with $k=50$), and nucleus sampling ($\eta=0.95$). The hyperparameter settings for these decoding strategies were based on recommendations from prior work~\cite{Holtzman2020The, shumailov2024ai, arias_decoding_2024}.

Figure~\ref{fig:fully_synthetic} depicts the perplexity and evaluation accuracy on the WikiText-2 test set for every model generation. Additionally, we obtain scores for the qualitative metrics using the outputs generated by the model before being used for training (i.e. $\{\hat{\mathbf{x}}\}_{s=1}^n$ of $\mathcal{D}_{\text{S}}^i$ in Algorithm~\ref{alg:recursive_training}), and enumerate them in Table~\ref{tab:data_quality_fully_synthetic} for generations $0$ and $9$.

We observe that deterministic decoding strategies lead to the most severe model collapse, with perplexity scores increasing by a factor of $\sim$$2.5$. Stochastic sampling methods exhibit linear degradation across generations, while collapse accelerates under greedy decoding and beam search before plateauing in later generations. Prior to recursive training (at generation $0$), deterministic strategies yield significantly less fluent and more repetitive text, with MAUVE scores $<\!5\%$ and diversity scores $<\!20\%$. The disparity in data quality between deterministic and stochastic strategies in the open-ended text generation task has been explored in related literature~\cite{Holtzman2020The,pillutla2021mauve}. Here, we demonstrate that this disparity compounds across recursive training, resulting in a significant drop in downstream performance and output quality at generation $9$. While deterministic methods are rarely used in open-ended generation, we included them to facilitate a better comparison with~\citet{shumailov2024ai} (where beam-search is used), but exclude them in subsequent experiments due to their unrealistic collapse.

\begin{figure*}[t]
\centering
\resizebox{0.8\textwidth}{!}{%
\begin{tikzpicture}[
    node distance=1.5cm and 1.8cm,
    auto,
    >={Stealth[length=2.5mm, width=2mm]},
    diam/.style={
        diamond, draw=black, thick,
        minimum size=1.2cm, font=\Large
    },
    circ/.style={
        circle, draw=black, thick,
        minimum size=1.2cm, font=\Large
    },
    link/.style={->, thick}
]

\node[diam] (DH) {$\mathcal{D}_{\text{H}}$};
\node[circ, right=of DH] (p0) {$p^0$};
\node[diam, right=of p0] (DS1) {$\mathcal{D}_{\text{S}}^1$};
\node[circ, right=of DS1] (p1) {$p^1$};
\node[right=of p1] (dots) {$\dots$};
\node[diam, right=of dots] (DSn) {$\mathcal{D}_{\text{S}}^n$};
\node[circ, right=of DSn] (pn) {$p^n$};

\draw[link] (DH) -- node[below] {Train} (p0);
\draw[link] (p0) -- node[below] {Generate} (DS1);
\draw[link] (p1) -- (dots);
\draw[link] (dots) -- (DSn);

\draw[link, color=green!50!black] (DS1) -- node[above] {$\alpha$} (p1);
\draw[link, color=green!50!black] (DSn) -- (pn);

\draw[link, color=red] (DH) to[bend left=20] node[above] {$\beta$} (p1);
\draw[link, color=red] (DH) to[bend left=25] (pn);

\draw[link, color=blue] (DS1) to[bend right=25] node[below] {$\gamma$} (pn);

\end{tikzpicture}}
\caption{\textbf{Recursive training process}. The human data, $\mathcal{D}_{\text{H}}$, is used to train the first model iteration, $p^0$, which is then prompted to generate a synthetic dataset, $\mathcal{D}_{\text{S}}^1$. Subsequently, depending on the mixing coefficients $\alpha$, $\beta$, and $\gamma$, further model iterations are trained on a mix of human and synthetic data from the previous generation(s).}
\label{fig:recursive_training_process}
\end{figure*}

The effect of sampling directly from the probability distribution varied significantly between the two models. For \texttt{GPT-2}, pure sampling produces diverse and fluent text at generation $0$, but training recursively on these outputs results in the worst test perplexity among stochastic methods ($58.64$), and generated text that has low similarity to human text (MAUVE of $7.18$). In contrast, with \texttt{SmolLM2}, pure sampling yields the smallest decline in task performance and maintains the closest overall similarity to human references across all evaluated metrics. Performance with top-$k$ sampling, on the other hand, was consistent across models. It led to the second-best task performance of all decoding strategies and, compared to the other truncation strategy, nucleus sampling, resulted in a smaller drop in diversity and stronger semantic resemblance to the human reference.

Temperature sampling led to the most repetitive outputs after recursive training, with diversity decreasing by $\sim$$70\%$ in both models. For \texttt{SmolLM2}, it also resulted in the greatest semantic divergence from human-generated text, indicating pronounced model collapse. While temperature is often deployed in open-ended text generation to improve data quality~\cite{Holtzman2020The}, our results demonstrate that recursively training on synthetic data generated with temperature sampling can lead to model output convergence and a dramatic reduction in diversity.

In our subsequent experiments on preventing model collapse, we seek to validate that our method can work in the most challenging yet realistic scenario. For this reason, we evaluate the models using the worst-performing stochastic decoding method: pure sampling for \texttt{GPT-2} and temperature sampling for \texttt{SmolLM2}. In addition, to facilitate direct comparisons, we also evaluate with top-$k$ decoding due to the consistent performance across models.

\section{Preventing model collapse}
\label{sec:preventing_collapse}

So far, we have carried out recursive training in a setting where models are trained exclusively on the outputs of the previous generation without implicitly including any human-generated samples. We now turn our focus to the partially synthetic setting, a more realistic scenario where human data make up a portion of the training dataset and the synthetic data is a mix of the samples produced across generations. The training dataset for generation $i$, $\mathcal{D}^{i}$, consists of the aggregation of $3$ samples:
\begin{equation}\label{eq:di}
\begin{split}
    \mathcal{D}^{i} \sim & \,\, {\text{\small sample}_{i \ge 1}} \,\, \mathcal{D}_{\text{H}},\alpha \\
                         & \,\, {\text{\small sample}_{i \ge 1}} \,\, \mathcal{D}_{\text{S}}^{i}, \beta\\
                         & \,\, {\text{\small sample}_{i \ge 2}} \,\, \left\{\mathcal{D}_{\text{S}}^{i-1},\dots,\mathcal{D}_{\text{S}}^{1}\right\}, \frac{\gamma}{(i-1)} \, ,
\end{split}
\end{equation}
where $\alpha$, $\beta$, and $\gamma\!\in\![0, 1]$ are mixing coefficients that affect the distribution of human and machine-generated data as well as the proportion of cross-generational data in the training set. 

The recursive training process is depicted in Figure~\ref{fig:recursive_training_process}. We explore the following settings: (i) fully synthetic $\left(\alpha\!=\!0,\beta\!=\!1,\gamma\!=\!0\right)$, where training data consists entirely of synthetic samples from the previous generation, (ii) partially synthetic $\left(\alpha\!>\!0, \beta\!=\!1, \gamma\!=\!0\right)$, where the same proportion of human data is added to the training data at every generation, and (iii) partially synthetic with synthetic data accumulated across generations $\left(\alpha\!>\!0, \beta\!>\!0, \gamma\!>\!0\right)$  as proposed in \cite{gerstgrasser_is_2024}.
We evaluate our method in the partially synthetic setting and vary the mixing coefficients $\alpha$, $\beta$, and $\gamma$. However, our method does not assume access to the values of the mixing coefficients, and hence the data distribution. To prevent model collapse when the origin of each training sample is unknown, we train a machine-generated text detector. The detector estimates the likelihood that a text sample is generated by a human (section~\ref{subsec:perf_detector}). We then use this information to conduct importance resampling on the training data (section~\ref{subsec:detector_sampling}) that ultimately mitigates model collapse.

\begin{figure*}[t]
    \centering
    \includegraphics[width=0.9\linewidth]{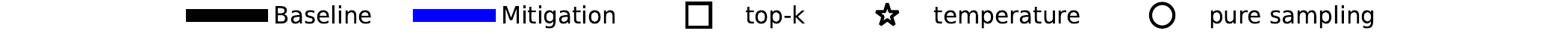}
    
    \begin{tabular}{@{}c@{\hspace{0.1cm}}c@{}c@{}c@{}c@{}c@{}}
        \raisebox{1.05cm}{\rotatebox{90}{\small\texttt{GPT-2}}} &
        \includegraphics[width=0.19\linewidth]{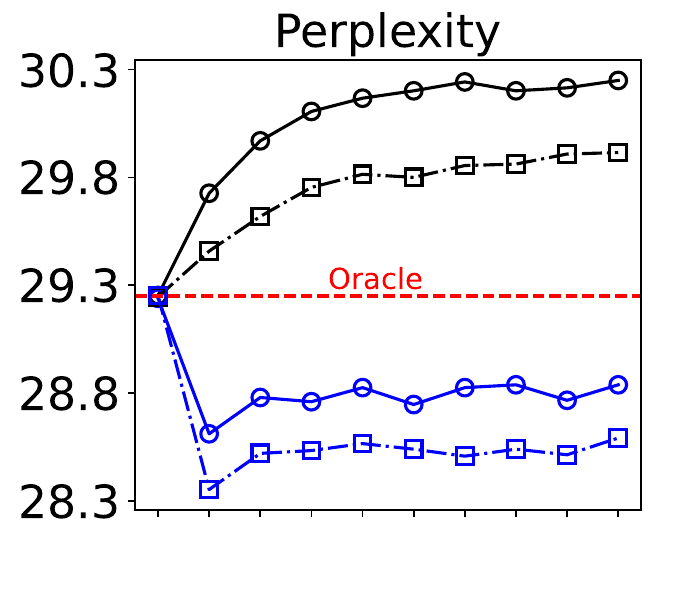} &
        \includegraphics[width=0.19\linewidth]{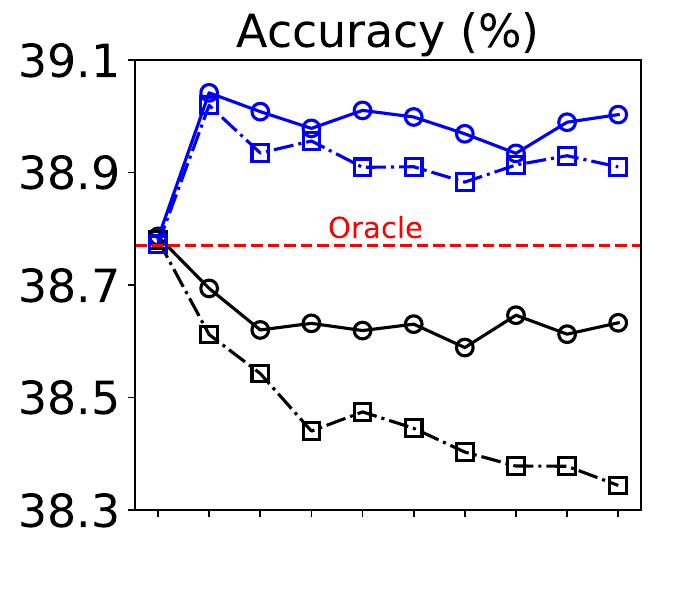} &
        \includegraphics[width=0.19\linewidth]{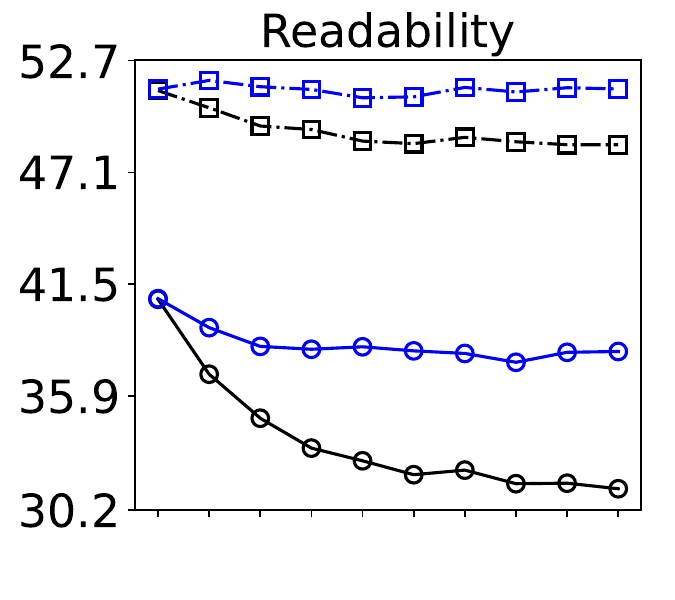} &
        \includegraphics[width=0.19\linewidth]{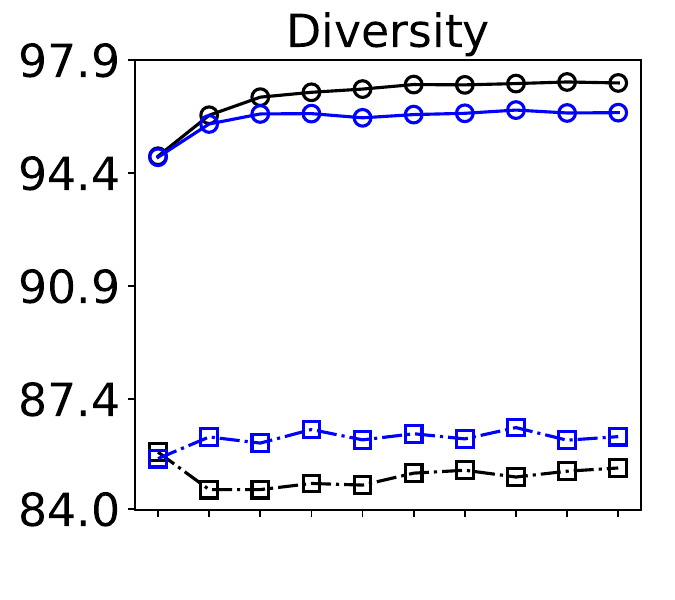} &
        \includegraphics[width=0.19\linewidth]{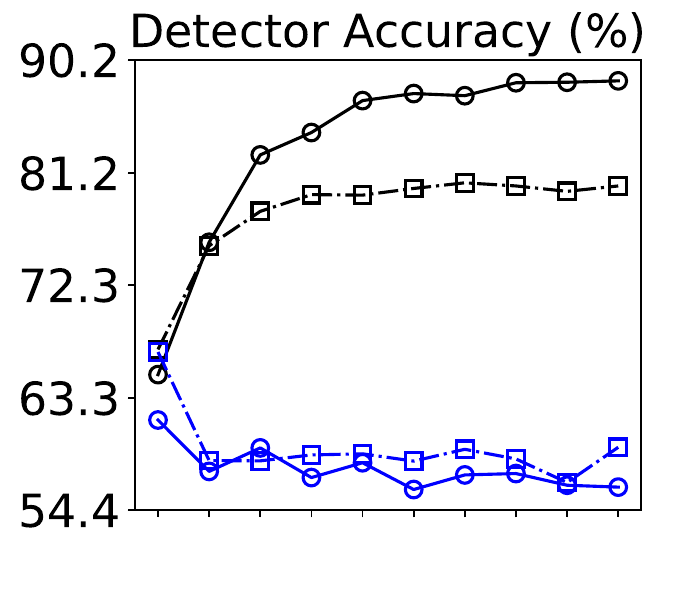} \\[-0.25cm]
        
        \raisebox{0.9cm}{\rotatebox{90}{\small\texttt{SmolLM2}}} &
        \includegraphics[width=0.19\linewidth]{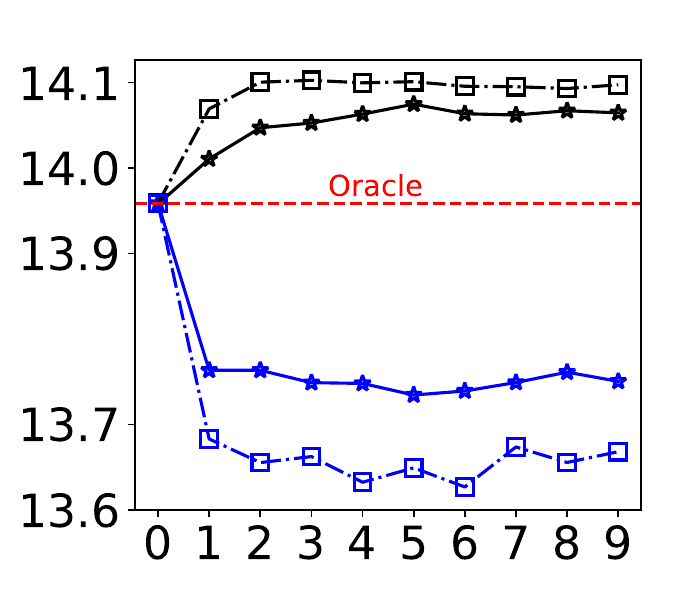} &
        \includegraphics[width=0.19\linewidth]{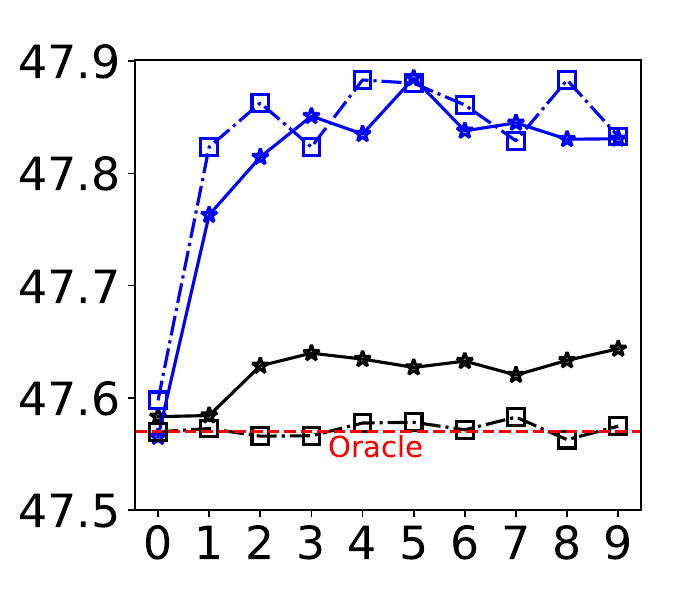} &
        \includegraphics[width=0.19\linewidth]{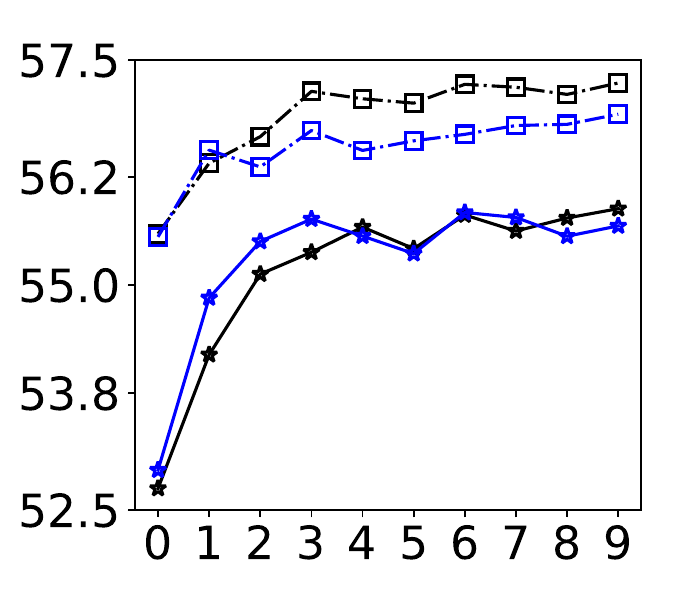} &
        \includegraphics[width=0.19\linewidth]{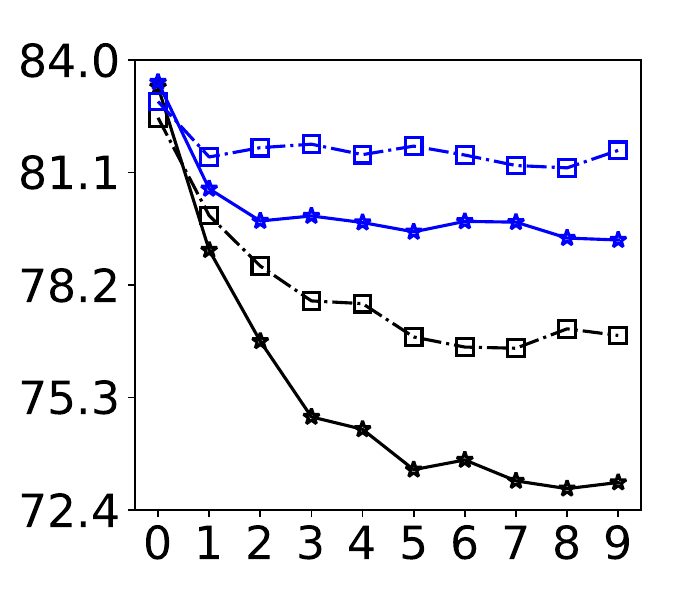} &
        \includegraphics[width=0.19\linewidth]{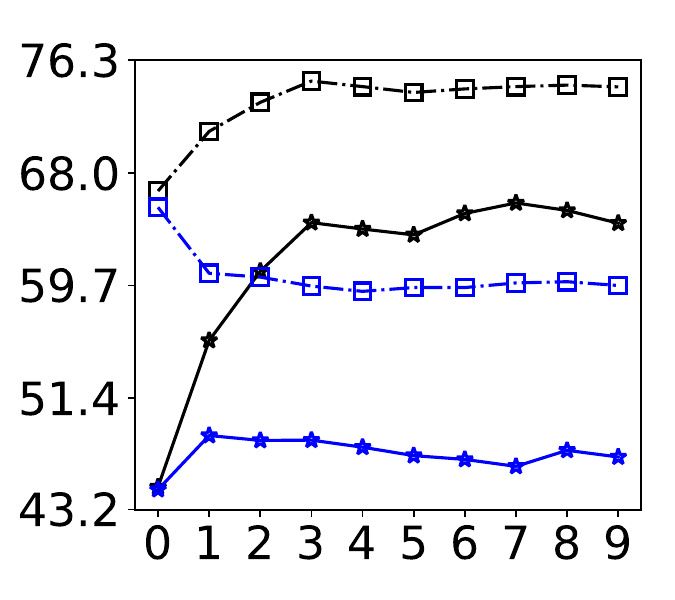} \\
    \end{tabular}
    
    \caption{Model collapse mitigation with \texttt{GPT-2} or \texttt{SmolLM2} under partially synthetic recursive training $\left(\alpha\!=\!1,\beta\!=\!1,\gamma\!=\!0\right)$ for generations $0$ to $9$. The baseline is equivalent to training on all the data in the pool and the `Oracle' performance represents a perfect machine-generated text detector that filters all synthetic samples.}
    \label{fig:partially_synthetic_a1}
\end{figure*}

\subsection{Machine-generated text detection performance}
\label{subsec:perf_detector}
For the machine-generated text detector, we trained and evaluated three pretrained models (\texttt{RoBERTa}, \texttt{DeBERTav3}, and \texttt{ModernBERT}) on the MAGE dataset \cite{mage}, a machine-generated text detection benchmark based on documents from $10$ domains. For each human-written text sample in the dataset, $27$ LLMs were prompted with the first $30$ words to generate a set of machine-generated samples. We adopt the preset training / validation / test splits ($80\%$/$10\%$/$10\%$). We also test on the more demanding out-of-distribution test set that contains human-curated text from $4$ unseen domains and machine-generated samples by an unseen LLM (\texttt{GPT-4}). Each model is fine-tuned for $5$ epochs using a binary cross-entropy loss and the best checkpoint is selected based on validation performance. More details on the model training can be found in Appendix~\ref{sec:ai_text_detection}. Performance is enumerated in Table~\ref{tab:ai_detectors}. \texttt{ModernBERT}\footnote{Our fine-tuned \texttt{ModernBERT} machine-generated text detector is available at \href{https://huggingface.co/GeorgeDrayson/modernbert-ai-detection}{\nolinkurl{huggingface.co/GeorgeDrayson/modernbert-ai-detection}}.} yielded the best classification performance on both test sets with an AUC of $.986$ and $.943$, respectively. This is comparable to the top-performing model evaluated by \citet{mage}, Longformer, which achieved an in-distribution and out-of-distribution AUC of $.99$ and $.94$, respectively.

\begin{table}[b]
\centering
\small
\setlength{\tabcolsep}{4pt}
\begin{tabular}{lcccccc}
\toprule
\multirow{2}{*}{\bf Model}& \multicolumn{3}{c}{\bf in-distribution} & \multicolumn{3}{c}{\bf out-of-distribution}\\
& AUC & Acc. & F1 & AUC & Acc. & F1 \\
\midrule
\texttt{RoBERTa}      & $.982$ & $.940$ & $.940$ & $.846$ & $.806$ & $.804$\\
\texttt{DeBERTav3}    & $.971$ & $\mathbf{.954}$ & $\mathbf{.954}$ & $.817$ & $.812$ & $.810$ \\
\texttt{ModernBERT}   & $\mathbf{.986}$ & $.948$ & $.948$ & $\mathbf{.943}$ & $\mathbf{.861}$ & $\mathbf{.860}$ \\
\bottomrule
\end{tabular}
\caption{Machine-generated text detection performance. Accuracy (Acc.) and F1-score are macro-averages.}
\label{tab:ai_detectors}
\end{table}

\subsection{Informed resampling of training data}
\label{subsec:detector_sampling}
Given a dataset at generation $i$, $\mathcal{D}_i \sim g(x)$, composed of an unknown mixture of human and synthetic samples, our goal is to sample a training dataset from a target human data distribution $\mathcal{D}_{\text{H}} \sim h(x)$ to prevent model collapse. We consider that a language model has collapsed when the inclusion of synthetic samples in the training data results in degraded performance compared to training exclusively on human samples.

\begin{table*}[t]
    \small
    \centering
    \setlength{\tabcolsep}{4pt}
    \resizebox{0.9\linewidth}{!}{%
    \begin{tabular}{@{}llccccccccc@{}}
        \toprule
        \bf Model & \bf Decoding & $\alpha$, $\beta$, $\gamma$ & \bf Perplexity$\downarrow$ & \bf Accuracy$\uparrow$ & \bf Diversity$\uparrow$ & \bf Self-BLEU$\downarrow$ & \bf MAUVE$\uparrow$ & \bf Readability$\uparrow$ \\
        \midrule
        \multirow{6}{*}{\texttt{GPT-2}} 
        & \multirow{3}{*}{top-$k$} 
            & $.5$, $1$, $0$   & $\textcolor{blue}{-7.28\%}$ & $\textcolor{blue}{+2.37\%}$ & $\textcolor{blue}{+4.54\%}$  & $\textcolor{blue}{-1.77\%}$  & $\textcolor{blue}{+0.46\%}$  & $\textcolor{blue}{+9.03\%}$  \\
        &   & $.5$, $.5$, $.5$ & $\textcolor{blue}{-5.94\%}$ & $\textcolor{blue}{+1.72\%}$ & $\textcolor{blue}{+2.43\%}$ & $\textcolor{blue}{-4.28\%}$ & $\textcolor{blue}{+3.99\%}$ & $\textcolor{blue}{+10.16\%}$  \\
        &   & $1$, $1$, $0$   & $\textcolor{blue}{-4.45\%}$ & $\textcolor{blue}{+1.49\%}$ & $\textcolor{blue}{+3.59\%}$  & $\textcolor{blue}{-3.58\%}$  & $\textcolor{blue}{+1.36\%}$  & $\textcolor{blue}{+6.76\%}$  \\
        \cmidrule{2-9}
        & \multirow{3}{*}{pure sampling}  
            & $.5$, $1$, $0$ & $\textcolor{blue}{-7.41\%}$ & $\textcolor{blue}{+1.50\%}$ & $\textcolor{red}{-1.30\%}$ & $\textcolor{red}{+36.71\%}$  & $\textcolor{blue}{+74.06\%}$  & $\textcolor{blue}{+50.23\%}$ \\
        &   & $.5$, $.5$, $.5$ & $\textcolor{blue}{-6.54\%}$ & $\textcolor{blue}{+1.50\%}$ & $\textcolor{red}{-1.07\%}$ & $\textcolor{red}{+21.39\%}$ & $\textcolor{blue}{+16.38\%}$ & $\textcolor{blue}{+25.45\%}$ \\
        &   & $1$, $1$, $0$ & $\textcolor{blue}{-25.65\%}$ & $\textcolor{blue}{+0.96\%}$ & $\textcolor{red}{-0.71\%}$ & $\textcolor{red}{+20.70\%}$ & $\textcolor{blue}{+16.42\%}$  & $\textcolor{blue}{+20.99\%}$  \\
        \midrule
        \multirow{6}{*}{\texttt{SmolLM2} $360$M}  
        & \multirow{3}{*}{top-$k$} 
            & $.5$, $1$, $0$ & $\textcolor{blue}{-4.60\%}$ & $\textcolor{blue}{+0.68\%}$ & $\textcolor{blue}{+7.06\%}$  & $\textcolor{blue}{-1.17\%}$ & $\textcolor{blue}{+15.28\%}$ & $\textcolor{red}{-0.73\%}$  \\
        &   & $.5$, $.5$, $.5$ & $\textcolor{blue}{-4.37\%}$ & $\textcolor{blue}{+0.62\%}$ & $\textcolor{blue}{+4.25\%}$ & $\textcolor{blue}{-0.42\%}$  & $\textcolor{blue}{+2.05\%}$  & $\textcolor{red}{-0.74\%}$\\
        &   & $1$, $1$, $0$ & $\textcolor{blue}{-3.05\%}$ & $\textcolor{blue}{+0.54\%}$ & $\textcolor{blue}{+6.20\%}$  & $\textcolor{blue}{-1.74\%}$ & $\textcolor{blue}{+10.62\%}$ & $\textcolor{red}{-0.61\%}$ \\
        \cmidrule{2-9}
        & \multirow{3}{*}{temperature}  
            & $.5$, $1$, $0$ & $\textcolor{blue}{-3.91\%}$ & $\textcolor{blue}{+0.55\%}$ & $\textcolor{blue}{+14.62\%}$  & $\textcolor{blue}{-1.46\%}$ & $\textcolor{blue}{+20.92\%}$ & $\textcolor{red}{-1.86\%}$  \\
        &   & $.5$, $.5$, $.5$ & $\textcolor{blue}{-3.24\%}$ & $\textcolor{blue}{+0.48\%}$ & $\textcolor{blue}{+7.44\%}$  & $\textcolor{blue}{-1.98\%}$  & $\textcolor{red}{-0.33\%}$  & $\textcolor{red}{-1.35\%}$ \\
        &   & $1$, $1$, $0$ & $\textcolor{blue}{-2.23\%}$ & $\textcolor{blue}{+0.39\%}$ & $\textcolor{blue}{+8.55\%}$ & $\textcolor{blue}{-1.80\%}$ & $\textcolor{blue}{+13.98\%}$  & $\textcolor{red}{-0.34\%}$ \\
        \midrule
        \texttt{SmolLM2} $135$M & top-$k$ & $1$, $1$, $0$ & $\textcolor{blue}{-4.19\%}$ & $\textcolor{blue}{+0.75\%}$ & $\textcolor{blue}{+9.96\%}$ & $\textcolor{blue}{-2.15\%}$ & $\textcolor{blue}{+13.08\%}$ & $\textcolor{red}{-0.76\%}$ \\
        \midrule
        \texttt{SmolLM2} $1.7$B & top-$k$ & $1$, $1$, $0$ & $\textcolor{blue}{-0.85\%}$ & $\textcolor{blue}{+0.06\%}$ & $\textcolor{blue}{+3.44\%}$ & $\textcolor{blue}{-1.11\%}$ & $\textcolor{blue}{+8.46\%}$ & $\textcolor{red}{-0.65\%}$ \\
        \bottomrule
    \end{tabular}
    }%
    \caption{Percentage of change in data quality and model performance when using our proposed mitigation strategy versus the baseline (final model generation). Results are shown for top-$k$ decoding and pure sampling/temperature for different values of $\alpha$, $\beta$, and $\gamma$ (\textcolor{blue}{blue} / \textcolor{red}{red}: \textcolor{blue}{positive} / \textcolor{red}{negative} results, $\uparrow$ / $\downarrow$: higher / lower is better).}
    \label{tab:data_quality_mitigation_diff}
\end{table*}

We use Sampling Importance Resampling (SIR) \cite{rubin1988}, a method for approximately sampling from a target distribution $h(x)$ based on sampling with importance weights from a proposal distribution $g(x)$ using the normalised likelihood ratio $h(x)/g(x)$. As this ratio is intractable in our case, we instead employ a machine-generated text detector to assign each sample, $\mathbf{x}_j$, a predicted probability $q\!\left(\mathbf{x}_j\right)$ of being machine-generated, which we treat as an approximation for the likelihood ratio.

As the detector has been trained on an unbalanced dataset ($29\%$ human samples), the predictions are biased towards attributing text as machine-generated, reflected in the optimal classification threshold of $0.8674$ ($0$/$1$: human/machine-generated). To ameliorate this, we apply a bias term $b\!\ge\!1$ (see Appendix~\ref{sec:informed_sampling_appendix}) to the probabilities, followed by normalising the weights using
\begin{equation}
   w_j = \frac{\left(1-q\!\left(\mathbf{x}_j\right)\right)^{b}}{\sum_{l=1}^{n} \left(1 - q\!\left(\mathbf{x}_l\right)\right)^b} \,\, ,
\end{equation}
where $w_j\!\in\![0,1]$ denotes the weight for sample $\mathbf{x}_j$, $\forall j$. From the $n$ weighted training samples, we draw $k\!\times\!n$ samples with replacement, with $k=1.5$ to allow for a $50\%$ upsampling of the training data. In this way, we obtain a revised set of samples that we use in our recursive training regime.

\subsection{Results on collapse prevention}
\label{subsec:collapse_prevention_results}

As explained in section~\ref{sec:exp_decoding}, we assess our approach by adopting the decoding strategy that caused the most significant model collapse, i.e. pure sampling for \texttt{GPT-2} and temperature sampling for \texttt{SmolLM2}. For a more direct comparison, we also conduct experiments using top-$k$ decoding for both models. At each generation $i$, we compare against the baseline of training on all samples in the pool of data $\mathcal{D}^i$ (Eq.~\ref{eq:di}). We also provide an `Oracle' performance, which internally deploys an infallible machine-generated text detector that filters all synthetic samples to ensure that the training dataset only contains human-written texts.

We evaluate recursive training in the partially synthetic setting under $3$ mixing settings: $\left(\alpha\!=\!1, \beta\!=\!1, \gamma\!=\!0\right)$, $\left(\alpha\!=\!0.5, \beta\!=\!1, \gamma\!=\!0\right)$, and $\left(\alpha\!=\!0.5, \beta\!=\!0.5, \gamma\!=\!0.5\right)$. The task performance, data quality, and detector accuracy metrics over $10$ generations of partially synthetic training are depicted in Figure~\ref{fig:partially_synthetic_a1} for the scenario where human and synthetic samples have equal proportion (Appendices~\ref{fig:partially_synthetic_a05}~and~\ref{fig:partially_synthetic_a05_g05} contain results for the other two scenarios).
The results show that our method of weighted resampling (section~\ref{subsec:detector_sampling}) prevents model collapse and preserves the readability and diversity of the synthetic outputs, while the baseline degrades in task performance and data quality. Over the generations, the outputs in the baseline become increasingly detectable, indicating a divergence from the characteristics of human-written text. In addition, our method improves performance compared to training exclusively on the human samples (`Oracle'), demonstrating both the value of using synthetic data in LLM training, but also the importance of selecting the right synthetic samples.

Table~\ref{tab:data_quality_mitigation_diff} enumerates the percentage of difference across various performance metrics for the final model generations under the baseline strategy vs. using our approach. We also enumerate the corresponding raw values in Table~\ref{tab:mitigation_percentages}. Our method improves the data quality and model performance across all metrics, except for pure sampling with \texttt{GPT-2}, where the baseline shows higher diversity and Self-BLEU but at the cost of lower MAUVE, readability, and task performance, indicating degraded model quality. Notably, in the setting where data is accumulated across generations $\left(\alpha\!=\!0.5, \beta\!=\!0.5, \gamma\!=\!0.5\right)$, we observe that mixing cross-generational data has a minimal effect on the extent of model collapse compared to training solely on the previous generation. This contrasts with the findings of~\citet{gerstgrasser_is_2024}, who suggest that accumulating all synthetic samples alongside the original human samples can avoid model
collapse. However, we note that~\citet{gerstgrasser_is_2024} did not constrain the sample size and trained on increasing volumes of data at each generation. Our experiments adopt a more realistic setting by sampling a fixed dataset size for each generation under different mixing scenarios.

\begin{figure*}[t]
    \centering
    \includegraphics[width=0.45\linewidth]{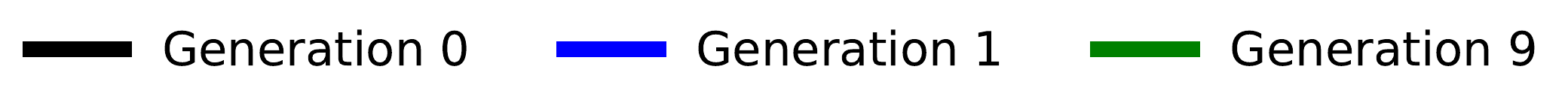}

    \begin{tabular}{@{}c@{\hspace{0.15cm}}c@{\hspace{0.15cm}}c@{\hspace{0.15cm}}c@{\hspace{0.15cm}}c@{}}
        & \multicolumn{2}{c}{\small\texttt{GPT-2}} 
        & \multicolumn{2}{c}{\small\texttt{SmolLM2}} \\

        \multirow{2}{*}[10.4ex]{\rotatebox{90}{\sffamily\small\bf Baseline}} &
        \includegraphics[width=0.21\linewidth]{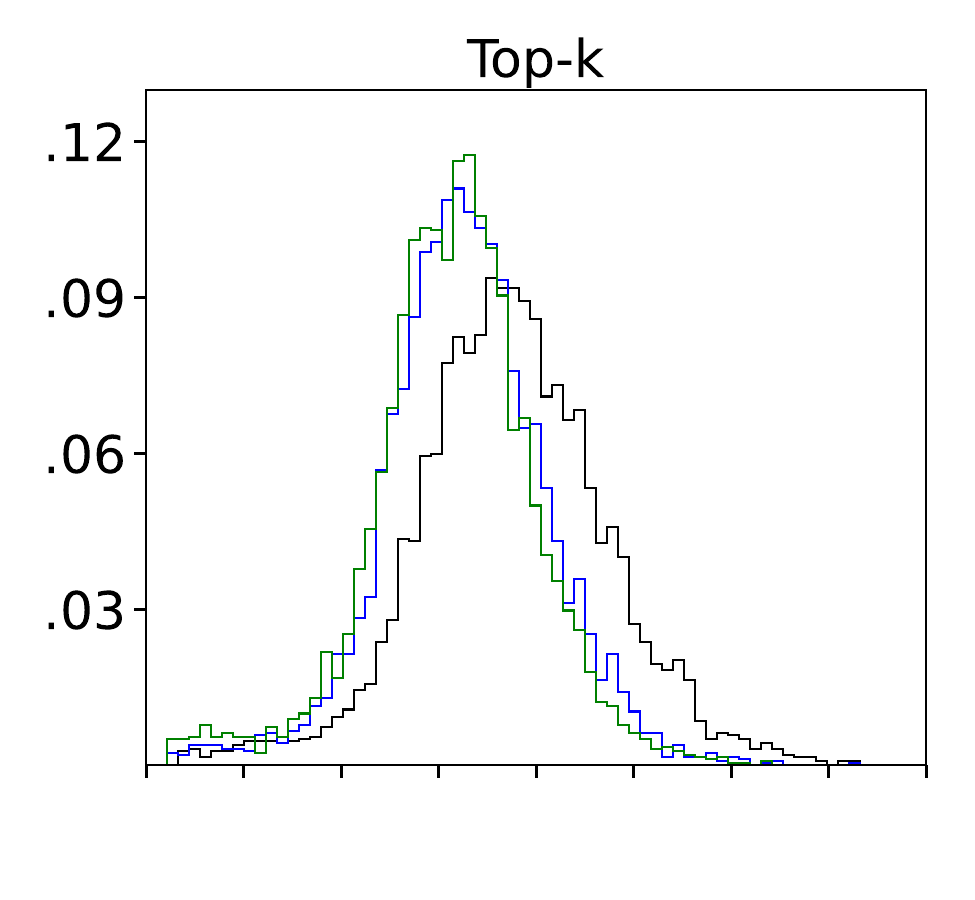} &
        \includegraphics[width=0.21\linewidth]{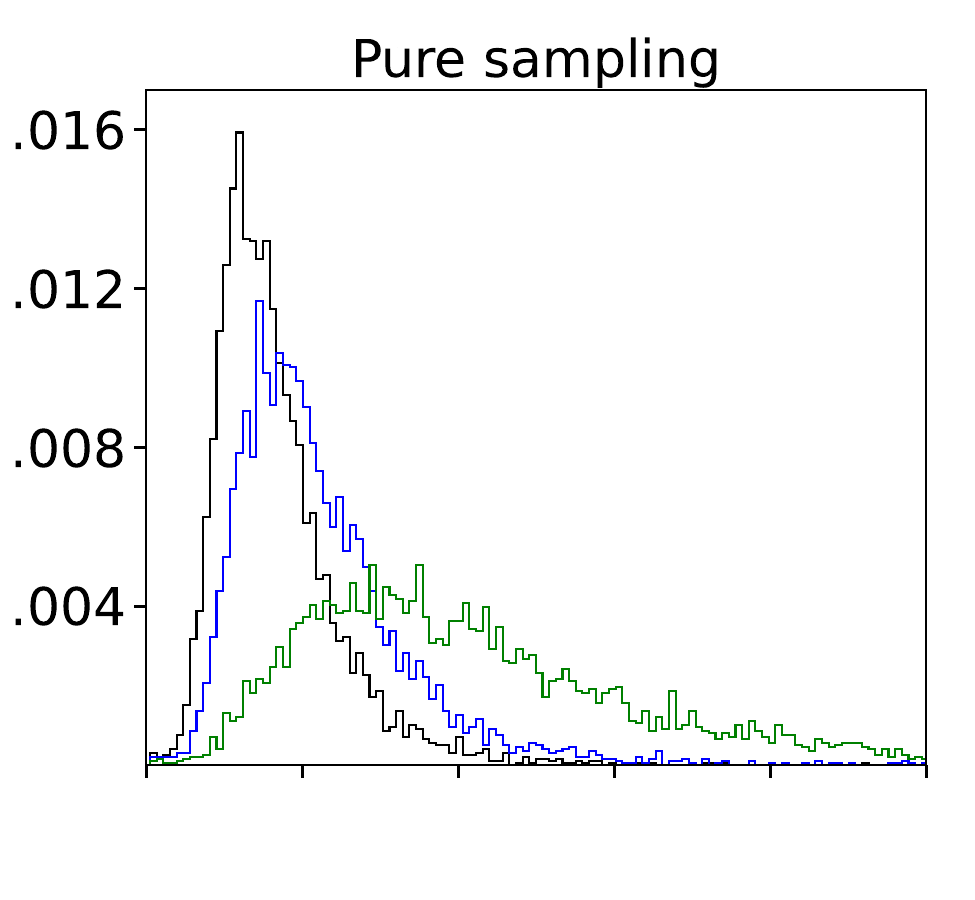} &
        \includegraphics[width=0.21\linewidth]{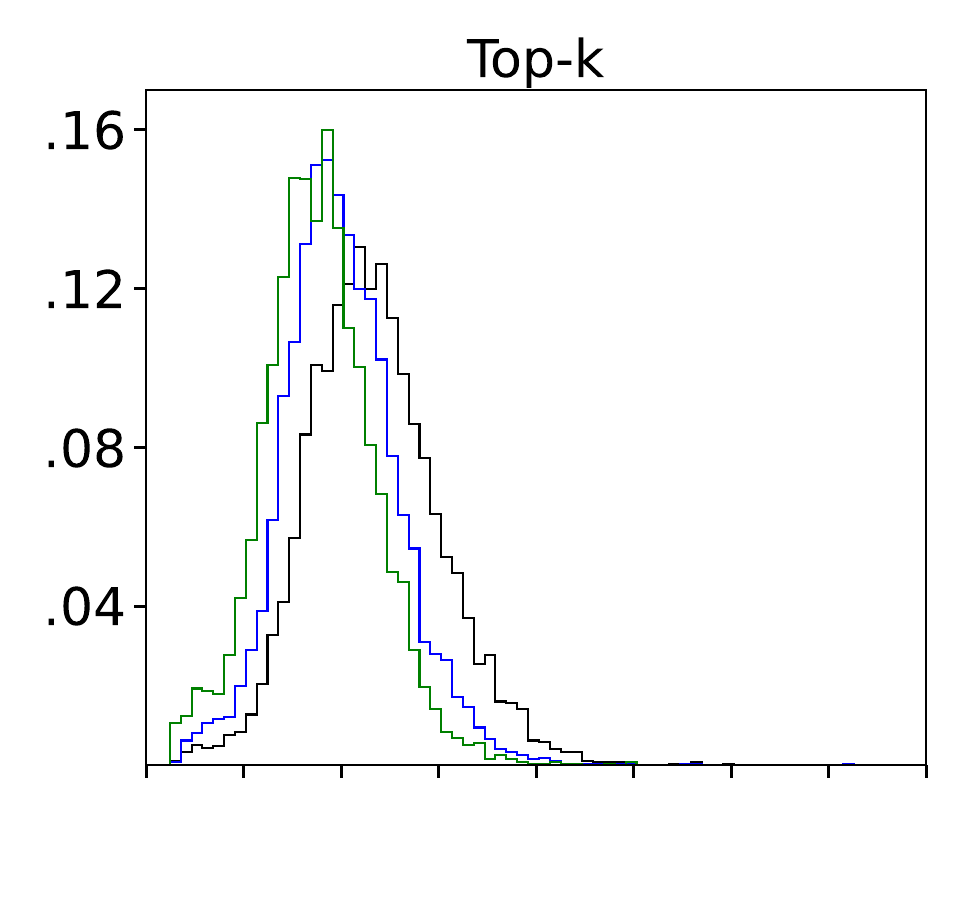} &
        \includegraphics[width=0.21\linewidth]{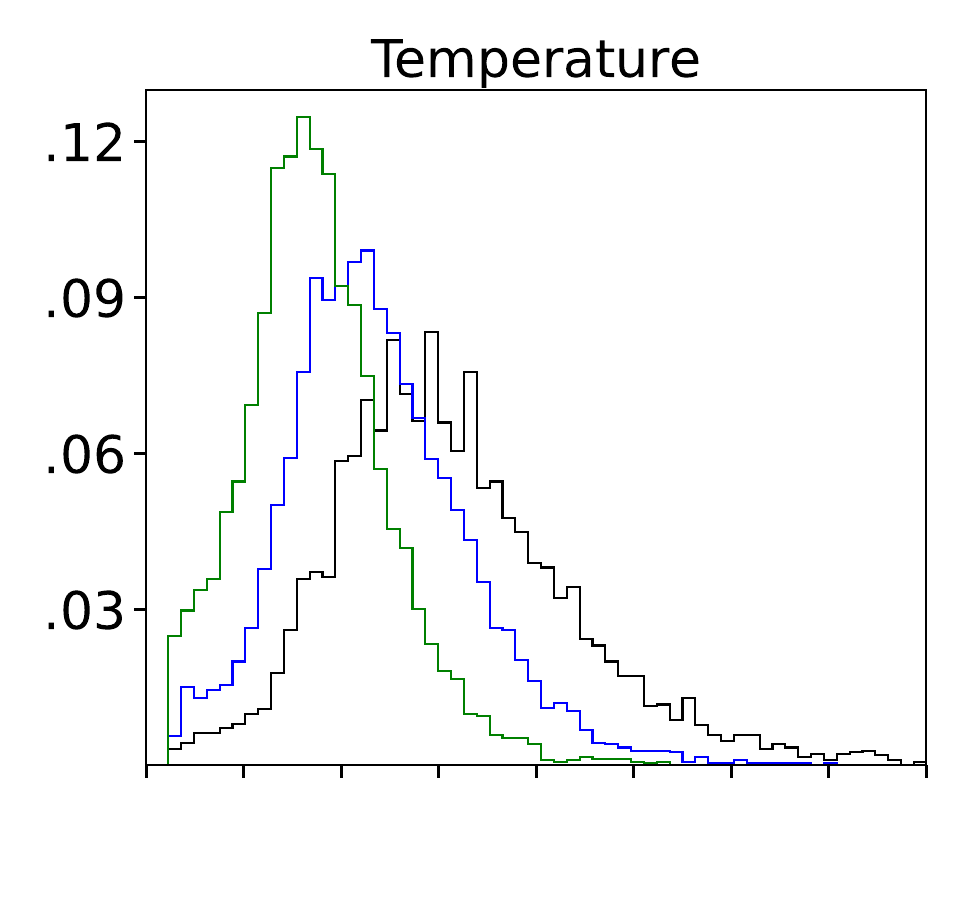} \\[-0.6cm]

        \multirow{2}{*}[11ex]{\rotatebox{90}{\sffamily\small\bf Mitigation}} &
        \includegraphics[width=0.21\linewidth]{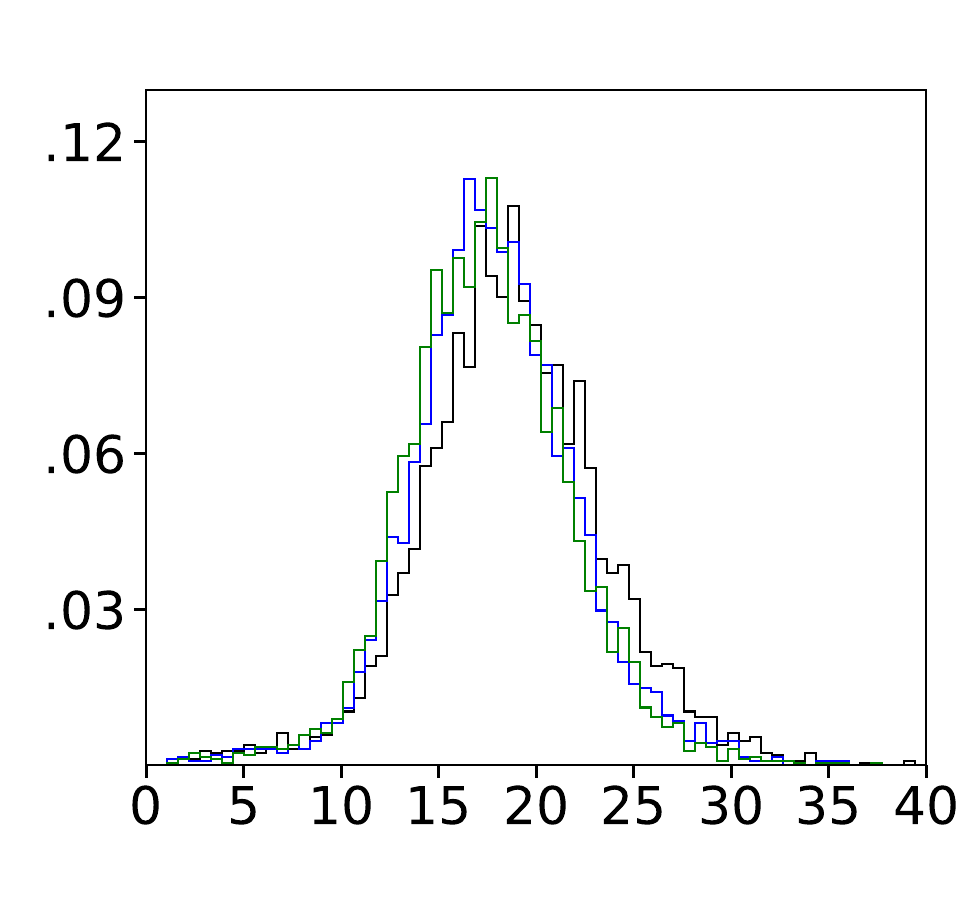} &
        \includegraphics[width=0.21\linewidth]{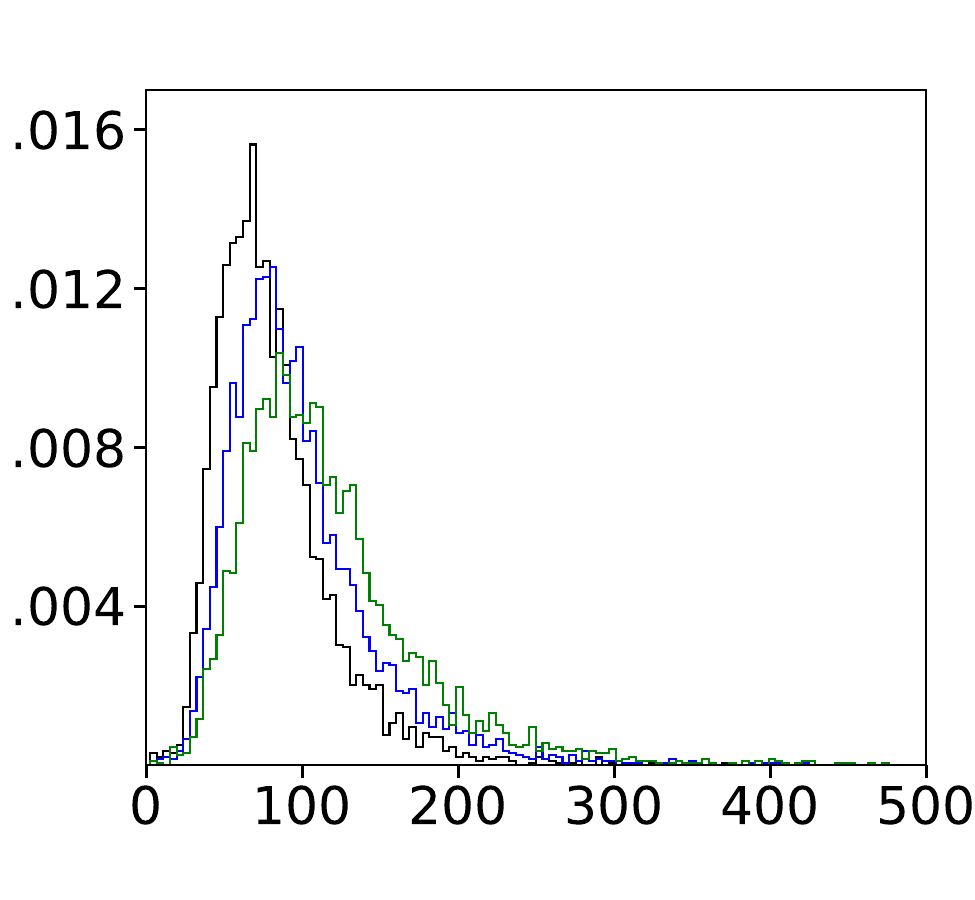} &
        \includegraphics[width=0.21\linewidth]{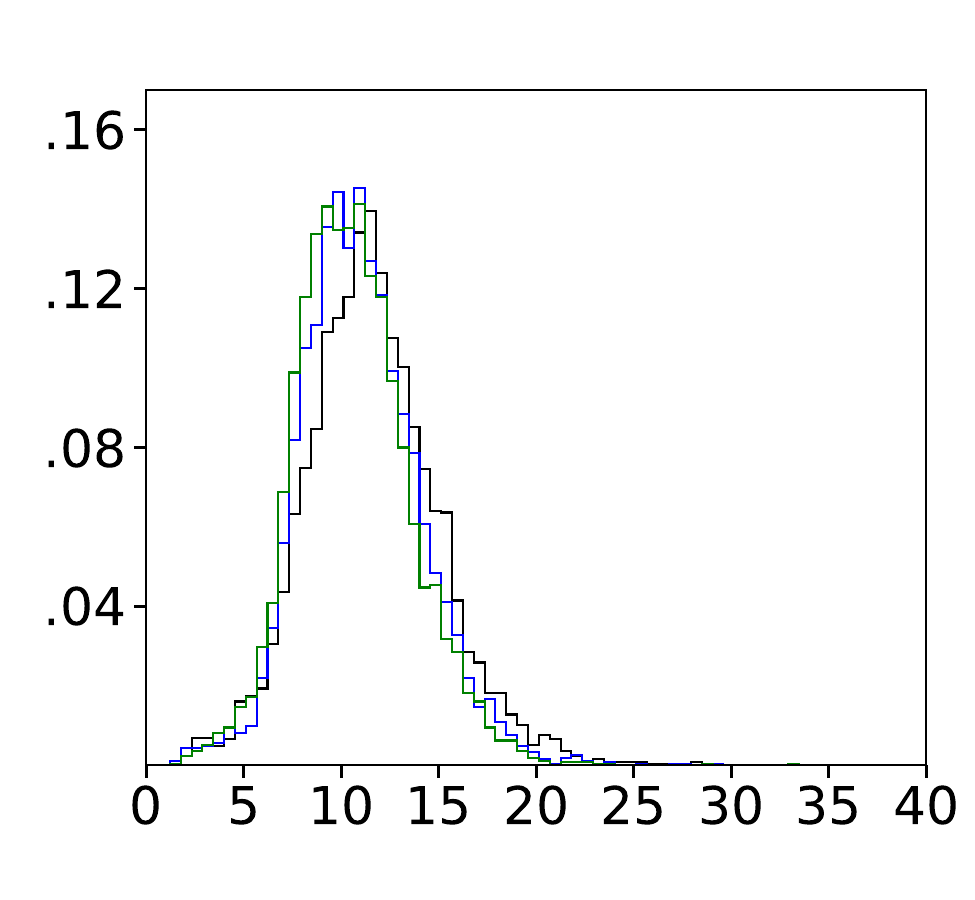} &
        \includegraphics[width=0.21\linewidth]{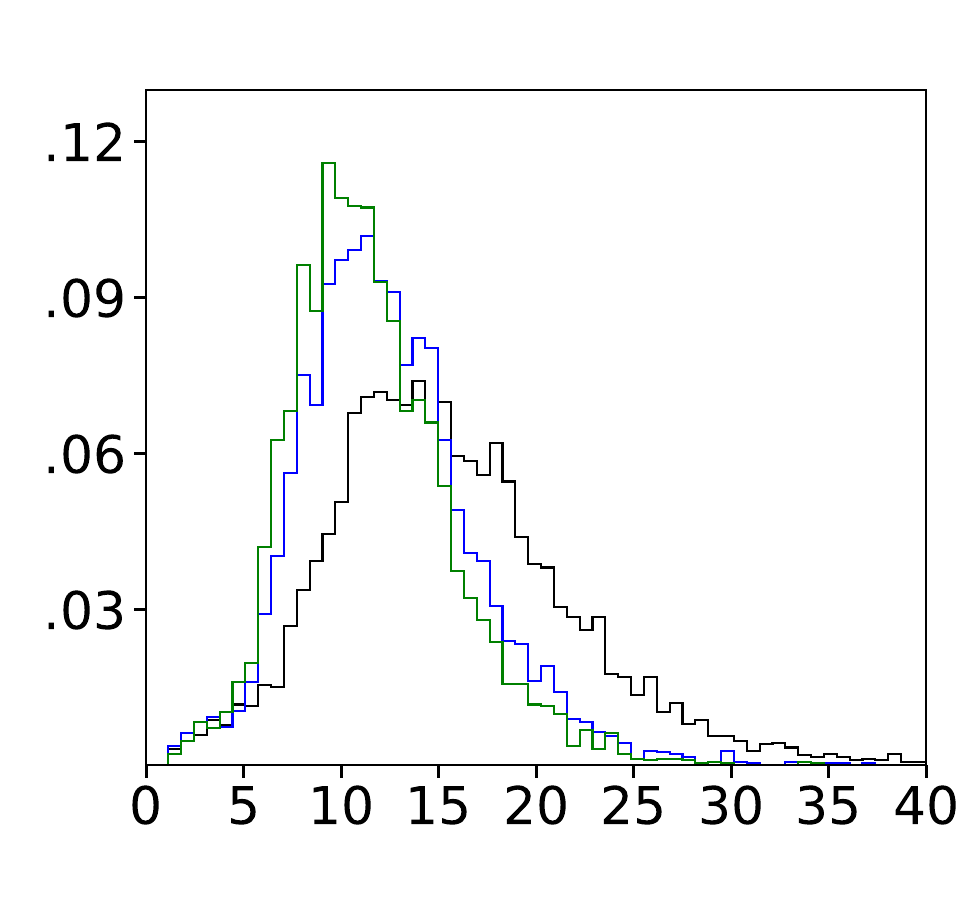} \\
    \end{tabular}

    \caption{Perplexity distribution of the machine-generated text at generations $0$, $1$, and $9$, evaluated using the model trained on human text ($p^0$). Results are shown for top-$k$, and pure sampling/temperature decoding for \texttt{GPT-2}/\texttt{SmolLM2} under the baseline strategy (top) and our importance resampling method (bottom) for the partially synthetic setting $\left(\alpha\!=\!0.5,\beta\!=\!1,\gamma\!=\!0\right)$.}
    \label{fig:perplexity_dist}
\end{figure*}

Additionally, we evaluate the effectiveness of our method at different model scales of \texttt{SmolLM2} ($135$M, $360$M, $1.7$B) using top-$k$ decoding due to its consistent effect across models and with a fixed data mixing setting $\left(\alpha\!=\!1,\beta\!=\!1,\gamma\!=\!0\right)$. The results are enumerated at the bottom of Table~\ref{tab:data_quality_mitigation_diff} and the performance across generations is depicted in Figure \ref{fig:partially_synthetic_model_size}. Our method prevents model collapse across all \texttt{SmolLM} model sizes, improving perplexity, accuracy, diversity, Self-BLEU, and MAUVE, with only minor reductions in readability. Notably, we find that smaller models exhibit greater relative improvements using our method compared to both the `Oracle' and the baseline, with the $135$M variant reducing perplexity by $4.19\%$ and increasing diversity by $9.96\%$. We hypothesise that this is because the data generated by smaller models is inherently of lower quality, leading to larger gains over the baseline by resampling the training data to reduce the proportion of synthetic samples.

In line with previous research~\cite{shumailov2024ai}, we also study the perplexity distribution of the synthetic data at each generation as evaluated by the original model $p^0$ trained on the human data. Figure~\ref{fig:perplexity_dist} depicts these distributions for generations $0$, $1$, and $9$ for \texttt{GPT-2} (top-$k$ and pure sampling decoding) and \texttt{SmolLM2} (top-$k$ and temperature sampling decoding) compared to the baseline. For top-$k$ decoding and temperature sampling, similarly to~\citet{shumailov2024ai}, we observe that over the generations, models produce samples that the original model, $p^0$, trained on the human data, would also be more likely to produce. This is depicted through perplexity scores, which shift towards regions of lower values, while their distribution becomes more peaked, showing signs of early model collapse. For pure sampling, on the other hand, we observe that the distribution shifts to higher perplexity scores and displays increased variance. This is an interesting finding that demonstrates that by removing truncation or temperature from the decoding strategy, the narrowing effect of model collapse is diminished, and instead, model collapse is reflected by long-tail incoherent text that is completely distinct from the original human samples. By deploying our mitigation strategy, however, we observe very little change in the perplexity distribution for all three sampling strategies.

\section{Conclusion}
This work investigates model collapse across three dimensions: model performance, data quality, and semantic similarity to human samples. Our analysis highlights that the extent of model collapse and the effect on the output distribution are influenced by the decoding strategy. Truncating can lead to peaked distributions and repetitive models, while pure sampling can result in high perplexity and verbose outputs with low resemblance to human text. Using the decoding strategies that resulted in the most extreme collapse, we evaluated the partially synthetic scenario, where human and LLM samples are mixed into the training set. We proposed a method to mitigate model collapse by resampling the training distribution with importance weights guided by the predictions of a machine-generated text detector. We validated our method on two language model families (\texttt{GPT-2} and \texttt{SmolLM2}) across a range of decoding strategies and model sizes ($124$M to $1.7$B), showing that model collapse can be prevented in all cases. Additionally, when the ratio of human to synthetic samples in the initial training pool is equal, our resampling method yields an improved performance compared to training exclusively on human data.

\section*{Limitations}
As in previous studies~\cite{shumailov2023curse, dohmatob2024tale}, we assess LLMs exclusively in a fine-tuning setting rather than pre-training from scratch. While pre-training experiments could provide deeper insights, the computational cost and complexity of training large-scale models from the ground up make such an approach impractical in our case. Nevertheless, given that model collapse has been primarily evaluated in LLMs from a fine-tuning setting, the conclusions made in this work still align with the current body of research.

In addition, our study focuses primarily on open-ended text generation tasks. While this is a crucial area for understanding model collapse, our findings may not fully generalise to other domains, such as structured prediction or code generation, where the impact of model collapse may manifest differently. Future work could explore whether our resampling method remains effective across these domains.

Lastly, as our model collapse experiments require full fine-tuning of a new model at each generation (for all $10$ generations), it becomes computationally expensive and time-intensive to evaluate large-scale models. On top of that, for the experiments we present in our paper, computational load increases significantly because we evaluate $6$ decoding strategies and $4$ data mixing settings. For this reason, the largest model evaluated is under $2$B parameters, which is at the limit of our available resources. Future work could extend this analysis to larger models to validate the generalisability of these findings with large-scale SOTA models.

\section*{Acknowledgments}
G. Drayson is funded by the EPSRC grant ``AI Centre for Doctoral Training in Foundational Artificial Intelligence'' (EP/S021566/1). The authors would like to thank the area chairs who reviewed this paper for their constructive feedback.

\bibliography{custom}

\balance
\clearpage
\nobalance

\appendix
\setcounter{table}{0}
\renewcommand{\thetable}{S\arabic{table}}%
\setcounter{figure}{0}
\renewcommand{\thefigure}{S\arabic{figure}}%
\setcounter{equation}{0}
\renewcommand{\theequation}{S\arabic{equation}}%

\section*{\Large\centering Appendix}

\section{Recursive training}

\subsection{Dataset}

All our model collapse experiments use the raw variant of the WikiText-2 dataset \cite{merity2016pointer}, which is licensed under the Creative Commons Attribution Share Alike Licence (CC BY-SA 4.0). We train models on the training set, consisting of $36{,}718$ documents, and evaluate on the test set of $4{,}358$ documents. The WikiText-2 dataset was extracted from the `Good' or `Featured' article criteria specified by editors on Wikipedia and only covers the English language.

\subsection{LLMs}

\texttt{GPT-2} (Generative Pre-trained Transformer 2;~\citealt{radford2019language}) is a decoder-only transformer-based language model that is available under the MIT licence. \texttt{GPT-2} demonstrated that large-scale language models could perform various language tasks without task-specific training. We use the base variant, which contains $124$M parameters. \texttt{SmolLM2}~\cite{smollm2} is a family of compact and efficient language models developed by Hugging Face, available in three sizes: $135$M, $360$M, and $1.7$B parameters and is available under the APACHE 2.0 licence. The majority of our experiments use the $360$M parameter variant unless specified otherwise.

\subsection{Hyperparameters}
\label{sec:model_collapse_training}
In our experiments, we conduct full fine-tuning using a learning rate of $5\!\times\!10^{-5}$, batch size of $8$ and a dropout rate of $0.1$. For the AdamW optimizer, we set $\beta_1 = 0.9$, $\beta_2 = 0.999$, and $\epsilon=10^{-8}$. Each model is trained for $1$ epoch with the hyperparameters fixed for all experiments. We conducted $10$ iterations of recursive training.

\section{Machine-generated text detection}
\label{sec:ai_text_detection}

\subsection{Pre-trained models}

Robustly Optimized \texttt{BERT} pre-training approach (\texttt{RoBERTa};~\citealt{liu2019-roberta}) improves on the pre-training phase of \texttt{BERT}~\cite{devlin-2019}, an encoder-only transformer model that leverages masked language models to enable pre-trained deep bidirectional representations. The \texttt{RoBERTa} model optimised the pre-training procedure for \texttt{BERT} by training the model for longer and on more data, changing the masking pattern, and removing the next sentence prediction objective. The model is licensed under the MIT licence. We use the base variant which has $125$M parameters.

Decoding-enhanced \texttt{BERT} with Disentangled Attention (\texttt{DeBERTav3};~\citealt{he2023debertav}) is a \texttt{BERT}-based encoder only model enhanced with disentangled attention. \texttt{DeBERTav3} improves on \texttt{DeBERTa} by using Enhanced Mask Decoding and an \texttt{ELECTRA}-style pre-training objective, Replaced Token Detection, instead of Masked Language Modelling. The model is licensed under the MIT licence. We use the base variant which contains $86$M backbone parameters with an embedding layer of $98$M parameters.

\texttt{ModernBERT}~\cite{warner_smarter_2024} is a recent addition to the encoder-only transformer models that has been designed to increase downstream performance and efficiency on GPUs, particularly in long-context scenarios due to its $8{,}192$ native sequence length. The model was trained on $2$ trillion tokens and improves on the original \texttt{BERT} architecture with rotary positional embeddings (RoPE), unpadding, GeGLU layers, and alternating local-global attention demonstrating SOTA performance amongst encoder models across a range of classification and retrieval tasks. The model is available under the APACHE 2.0 licence. We conduct experiments with the base variant, which contains $150$M parameters.

\subsection{Hyperparameters}
\label{sec:ai_hyperparameters}

Each model is fine-tuned for $5$ epochs. We select the best model based on the highest AUC on the validation set. Optimisation is performed using AdamW with $\beta_1=0.9$, $\beta_2=0.98$, $\epsilon=10^{-6}$, and a weight decay of $10^{-2}$. These parameters were chosen based on prior work~\cite{warner_smarter_2024}. The label smoothing parameter $\alpha$ is set to 0.1, the seed was fixed at $42$, and the training batch size to $16$. The learning rate is set based on a hyperparameter sweep over [$1$, $1.5$, $2$, $3$, $4$]$\times 10^{-5}$. For \texttt{ModernBERT} the best-performing learning rate was $10^{-5}$. We implement temperature scaling by learning the temperature parameter using L-BFGS optimisation on the validation set. This is run for $50$ iterations with a learning rate of $0.01$.

\subsection{Dataset}
\label{sec:ai_dataset}

We trained and evaluated the machine-generated text detectors on the MAGE dataset~\cite{mage}, which is based on documents from $10$ domains: opinion statements (CMV \& Yelp reviews dataset), news articles (XSum \& TLDR dataset), question answering (ELI5), story generation (Reddit WritingPrompts \& ROC), commonsense reasoning (HellaSwag), knowledge illustration (SQuAD), and Scientific writing (SciGen). The authors sampled $1{,}000$ texts from each domain (apart from opinion statements and news articles with $804$ and $777$ samples respectively) and generated text using $27$ LLMs from $7$ model families, which include OpenAI, LLaMA, GLM, FLAN-T5, OPT, BigScience, and EleutherAI and their dataset is available under the APACHE 2.0 licence. For each human-written sample in the dataset, they generate a machine-generated version by providing the first $30$ tokens of human-written text as context to the LLM. In addition, for the OpenAI models, they implemented two other prompt strategies for relevant domains: `topical' prompts such as an argument or news title and `specified' prompts which contain information about the domain source. This results in $33{,}000$ ($=27{,}000+3\!\times\!2\!\times\!1{,}000$) machine-generated samples per source before processing and filtering. The authors split the dataset into train, validation and test splits in the ratio $80$:$10$:$10$. To mitigate data imbalance in the validation and test sets they sample additional human data from each data source. The resulting test set contains $28{,}741$ human, and $28{,}078$ machine-generated samples ($49\%$ machine-generated). The training set, however, is $71\%$ machine-generated. The total dataset consists of $154{,}078$ human-written and $294{,}381$ machine-generated texts. In addition to the previously described test set, we also evaluate our detector on their more challenging test set containing text from four unseen domains (CNN/DailyMail, DialogSum, PubMedQA, IMDB) with machine-generated text from an unseen model (\texttt{GPT-4}). This out-of-distribution test set contains $762$ human and $800$ machine-generated samples.

When evaluating the \texttt{ModernBERT} model fine-tuned on MAGE on the \texttt{SmolLM2} models, we observed a drop in detection performance compared to \texttt{GPT-2} text, with large variability across decoding strategies and model size. For \texttt{SmolLM2} $360$M the detector achieved a classification accuracy of $.601$ for top-$k$ decoding while for temperature sampling the accuracy was $.399$. To ameliorate this, we finetuned a new \texttt{ModernBERT} model on a larger corpus, containing the MAGE dataset and a subset of the RAID dataset \cite{raiddugan} for the \texttt{SmolLM2} models. The RAID dataset is the largest machine-generated text detection dataset including text samples generated by $11$ LLMs with $4$ decoding strategies, and spans text across $8$ domains. Additionally, RAID includes $11$ types of adversarial attacks, such as homoglyph substitutions, number insertions, article deletions, and paraphrasing and is licensed under the MIT licence. We partitioned the dataset into training, validation, and test splits in the ratio $80$:$10$:$10$, ensuring no cross-contamination of text segments generated from the same source document across splits. We balanced each split so that it contained an equal number of human and machine samples, stratified across model, decoding strategy, source domain, and adversarial attack (the whitespace and paragraph attacks were included). This resulted in balanced train, validation, and test splits comprising $128{,}352$, $16{,}044$, and $16{,}056$ samples, respectively.

\section{Informed sampling of training data}
\label{sec:informed_sampling_appendix}

As we perform sampling with replacement, there is the risk of excessive duplication of high-weight samples. To account for this, we introduce a maximum resample count parameter, $r_{\text{max}}$, which limits the number of times any individual sample can be selected. This constraint ensures diversity in the resampled dataset and prevents a small subset of high-weight samples from dominating the training distribution. To further correct for the classifier’s bias toward labelling samples as machine-generated, we introduce a bias term $b\!\ge 1$ to adjust the weight distribution. This formulation increases the selection probability of samples likely to be human, counteracting the bias introduced by the classifier’s skewed confidence distribution. We select values for $r_{\text{max}}$ and $b$ by evaluating each model on the Wikitext-2 validation set after 1 generation of recursive training. The optimal hyperparameters for each model configuration are reported in Table~\ref{tab:importance_sampling_params}.

\begin{table}[t]
    \centering
    \setlength{\tabcolsep}{6pt}
    \begin{tabular}{@{}lcc@{}}
        \toprule
        \bf Model & $r_{\text{max}}$ & $b$ \\
        \midrule
        \texttt{GPT-2}         & $10$ & $10$ \\
        \texttt{SmolLM $135$M}   & $10$ & $10$ \\
        \texttt{SmolLM $360$M}   & $5$  & $10$ \\
        \texttt{SmolLM $1.7$B}   & $3$  & $1$  \\
        \bottomrule
    \end{tabular}
    \caption{Optimal hyperparameters for Sampling Importance Resampling across different model scales using top-$k$ decoding.}
    \label{tab:importance_sampling_params}
\end{table}

\begin{figure*}[t]
    \centering
    \includegraphics[width=0.6\linewidth]{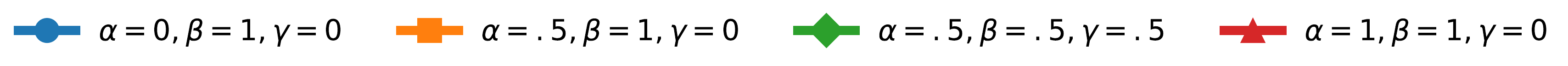}
    \begin{tabular}{c@{\hspace{0.25cm}}c@{\hspace{0.75cm}}c@{\hspace{0.25cm}}c@{}}
        \includegraphics[width=0.2\linewidth]{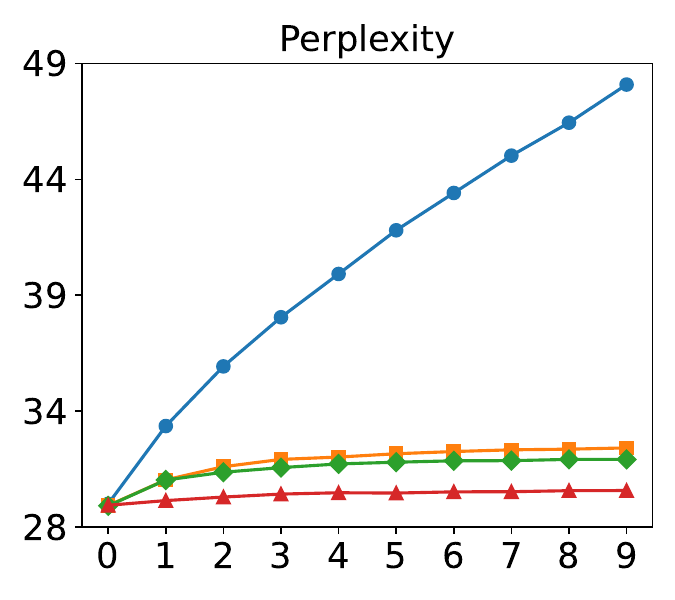} &
        \includegraphics[width=0.2\linewidth]{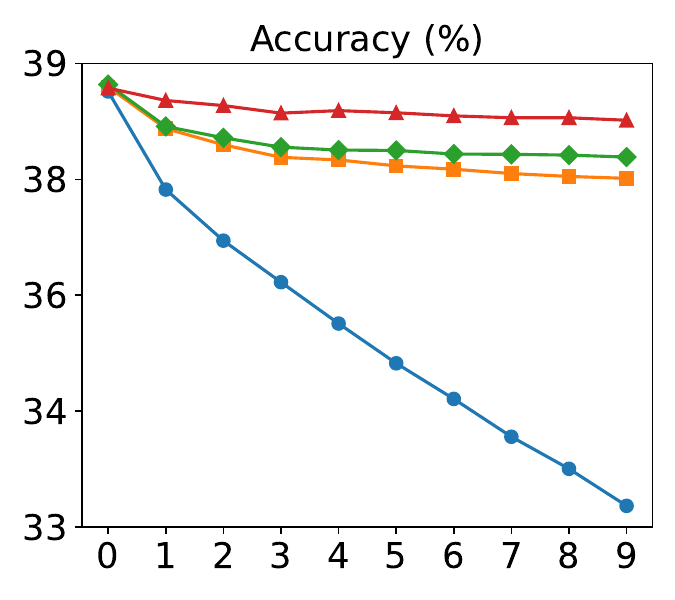} &
        \includegraphics[width=0.2\linewidth]{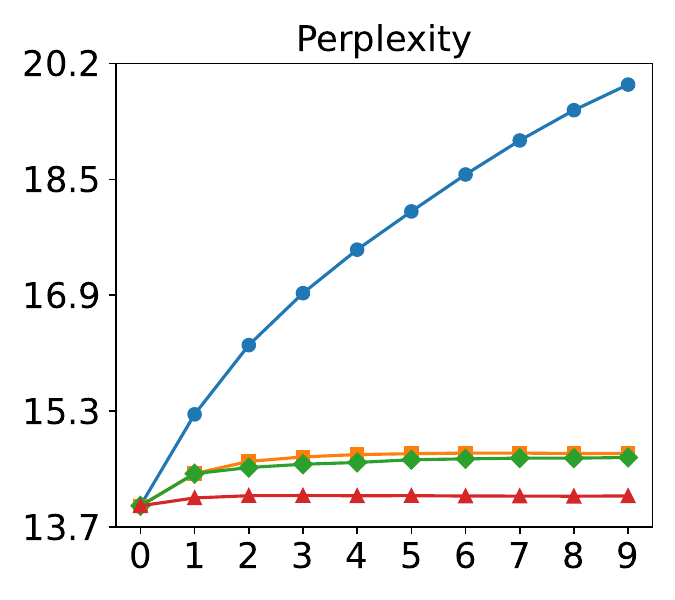} &
        \includegraphics[width=0.2\linewidth]{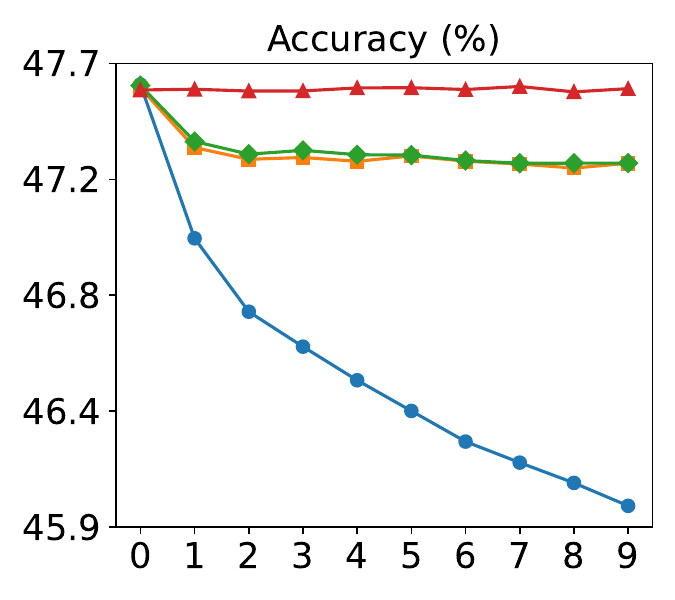}\\
        \multicolumn{2}{c}{\scriptsize\texttt{GPT-2}} & \multicolumn{2}{c}{\scriptsize\texttt{SmolLM2}}
    \end{tabular}
    \caption{Perplexity and accuracy over generations $0$ to $9$ of fully synthetic recursive training for varying mixing coefficients $\left(\alpha,\beta,\gamma\right)$ using top-$k$ decoding.}
    \label{fig:vary_alpha}
\end{figure*}

\begin{figure*}[t]
    \centering
    \includegraphics[width=0.3\linewidth]{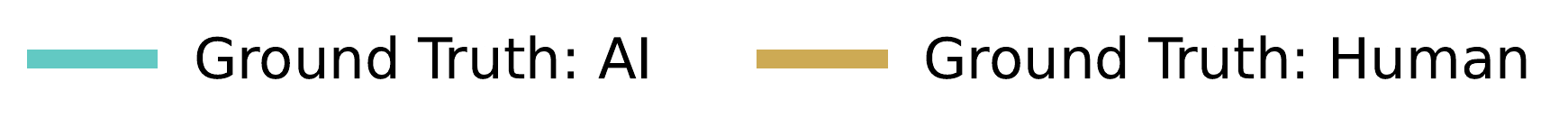}
    \begin{tabular}{c@{}c@{\hspace{0.5cm}}c@{}c@{\hspace{0.2cm}}}
        \includegraphics[width=0.22\linewidth]{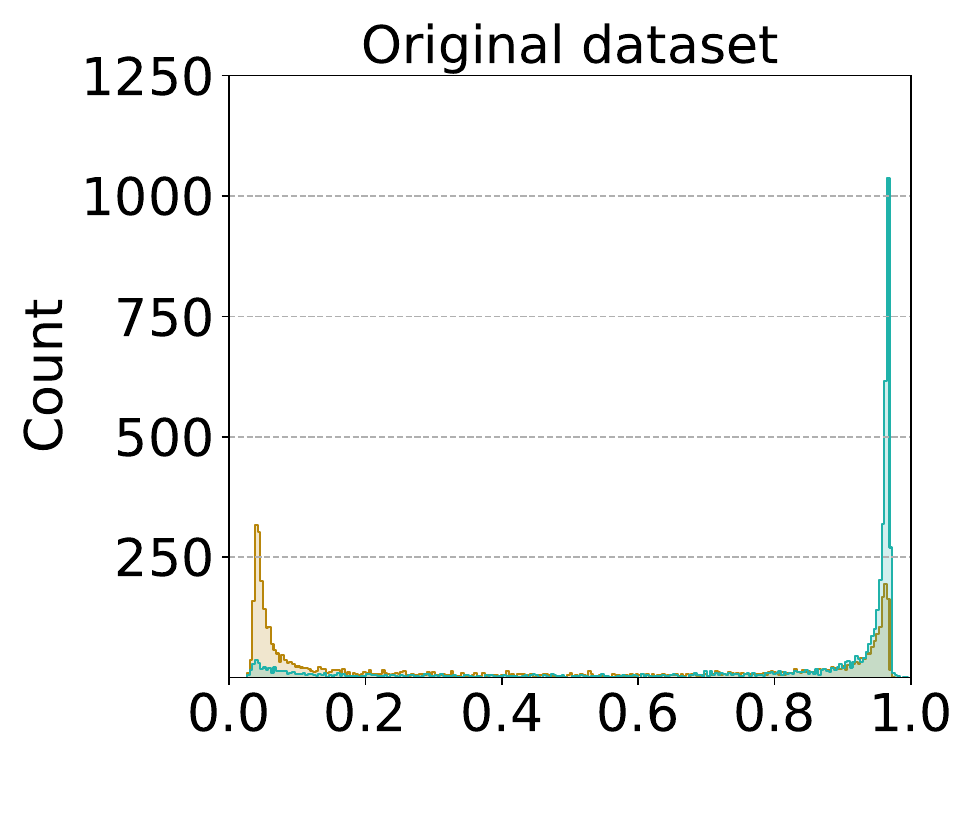} &
        \includegraphics[width=0.22\linewidth]{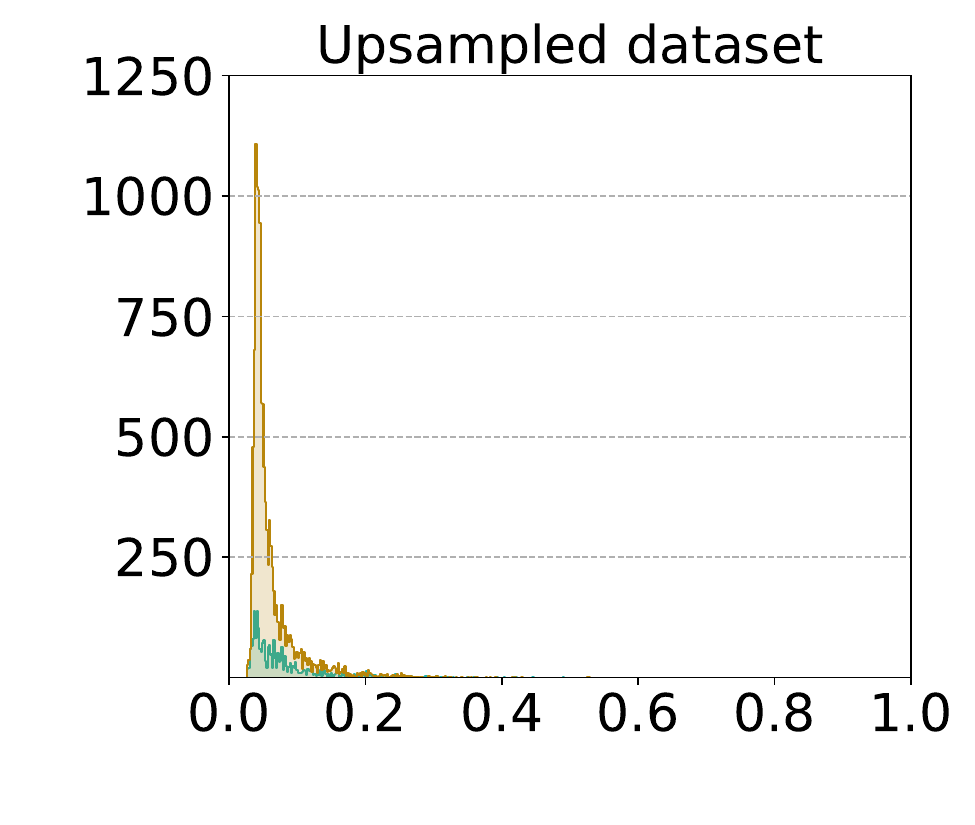} &
        \includegraphics[width=0.22\linewidth]{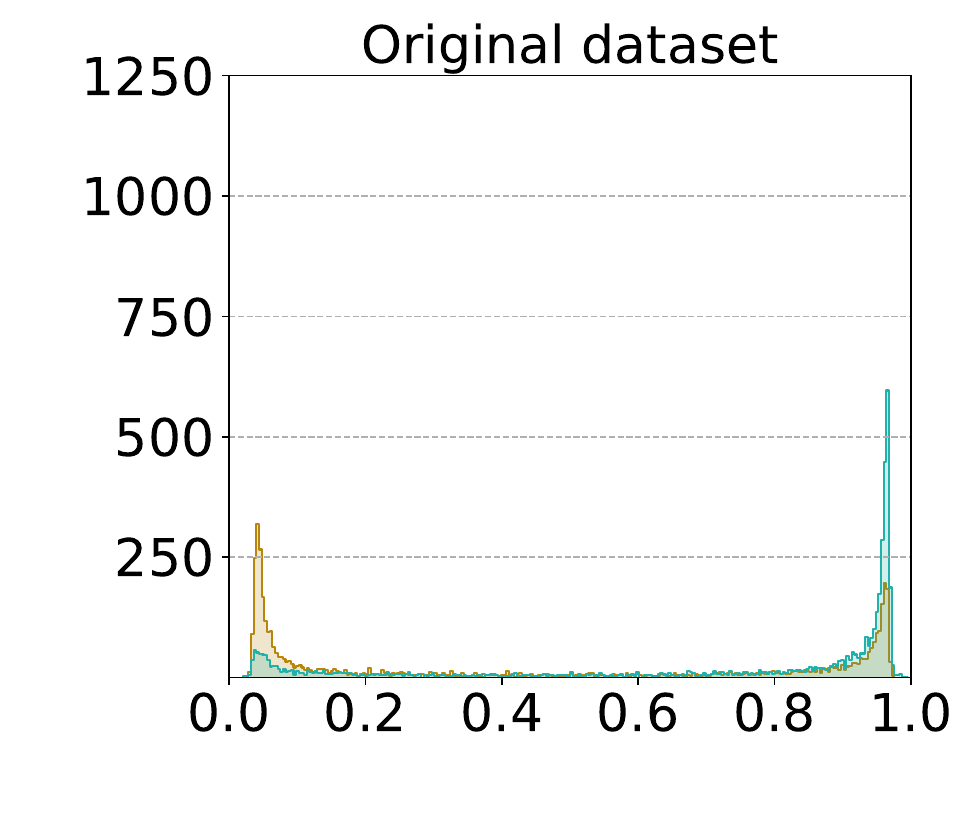} &
        \includegraphics[width=0.22\linewidth]{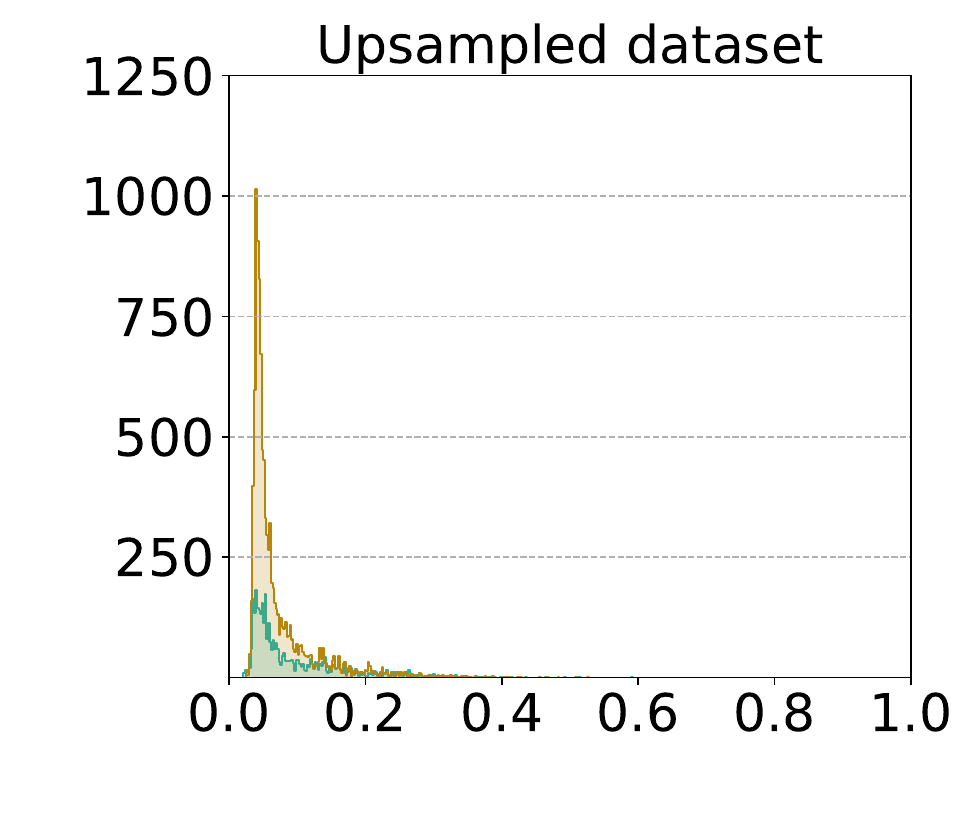}\\
        \multicolumn{2}{c}{\scriptsize\texttt{GPT-2}} & \multicolumn{2}{c}{\scriptsize\texttt{SmolLM2}}
    \end{tabular}
    \caption{Classification score distribution of the machine-generated text detector on the dataset from generation $0$.}
    \label{fig:confidence_distributions}
\end{figure*}

\begin{figure*}[t]
    \centering
    \includegraphics[width=0.85\linewidth]{images/legend.pdf}
    
    \begin{tabular}{@{}c@{\hspace{0.1cm}}c@{}c@{}c@{}c@{}c@{}}
        \raisebox{1.05cm}{\rotatebox{90}{\small \texttt{GPT-2}}} &
        \includegraphics[width=0.19\linewidth]{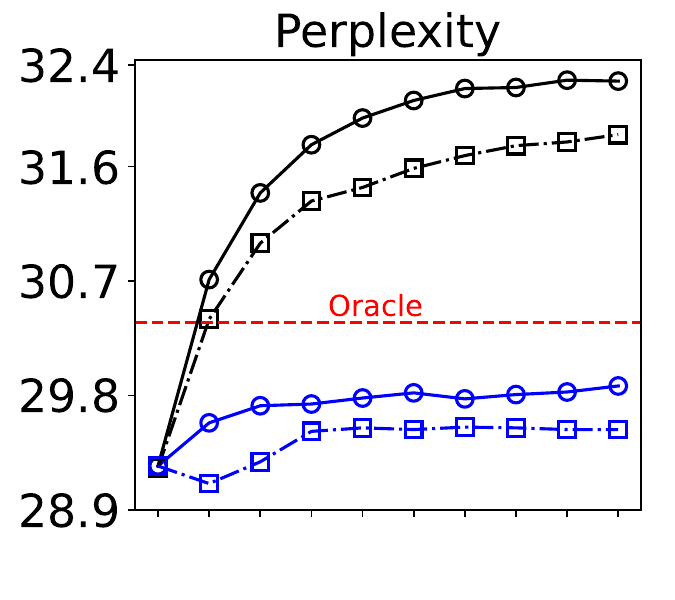} &
        \includegraphics[width=0.19\linewidth]{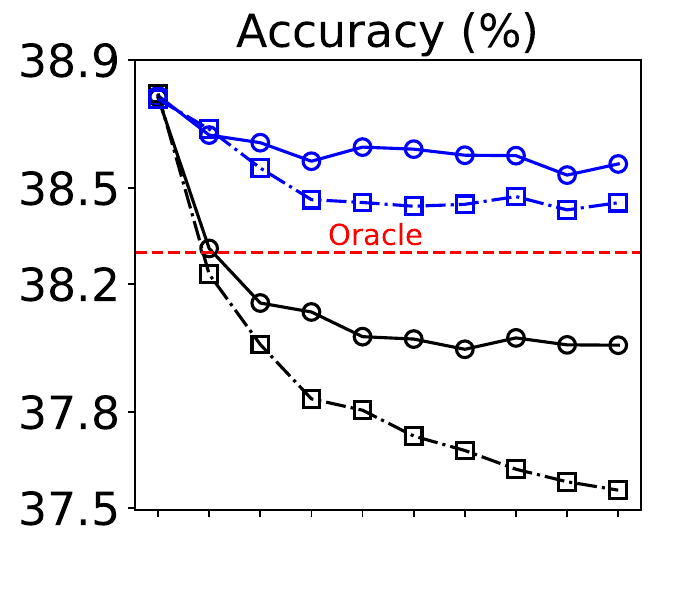} &
        \includegraphics[width=0.19\linewidth]{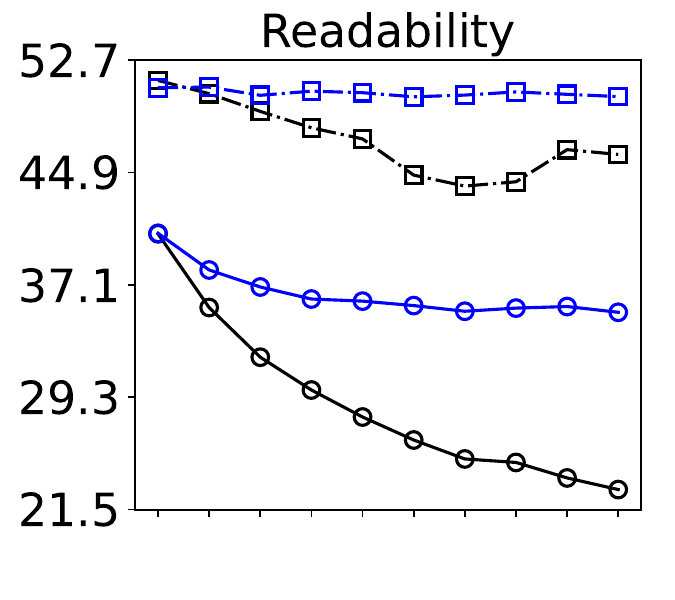} &
        \includegraphics[width=0.19\linewidth]{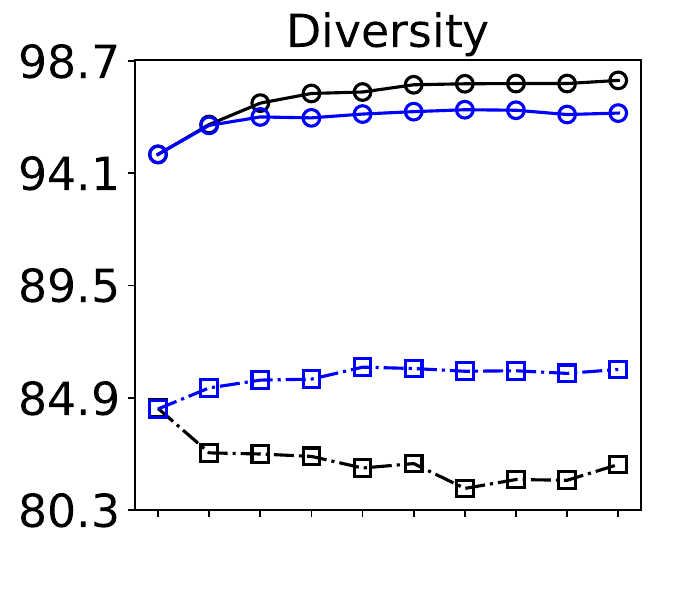} &
        \includegraphics[width=0.19\linewidth]{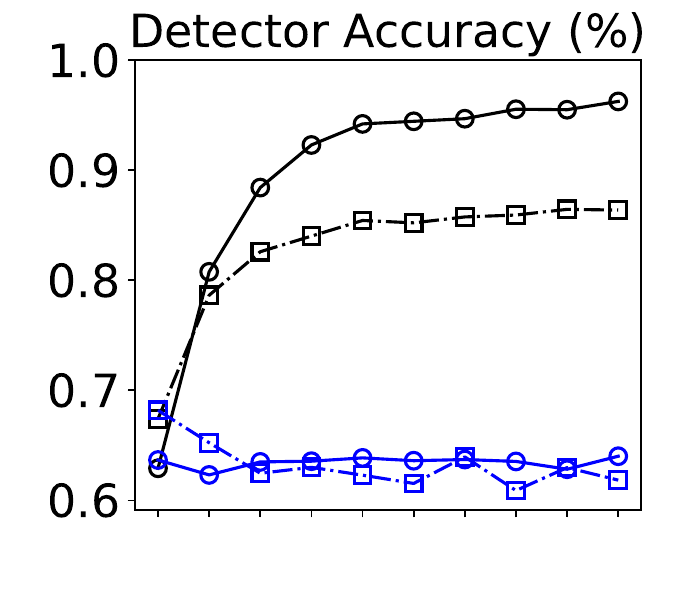} \\[-0.25cm]
        
        \raisebox{0.9cm}{\rotatebox{90}{\small \texttt{SmolLM2}}} &
        \includegraphics[width=0.19\linewidth]{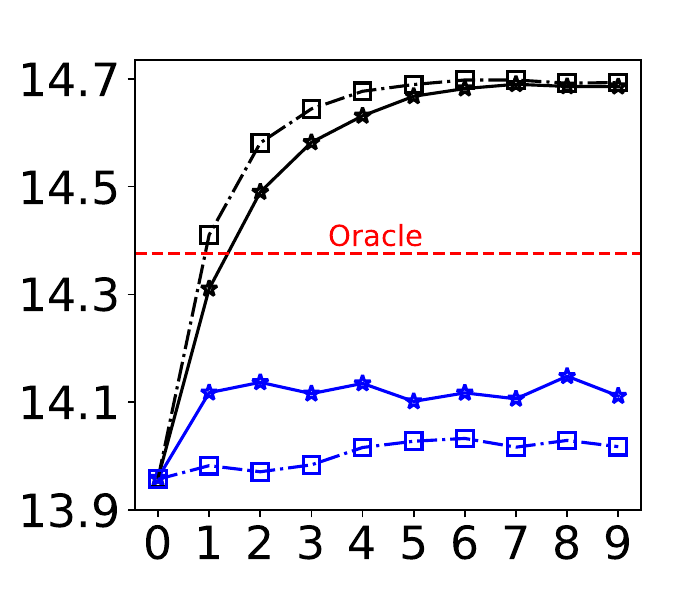} &
        \includegraphics[width=0.19\linewidth]{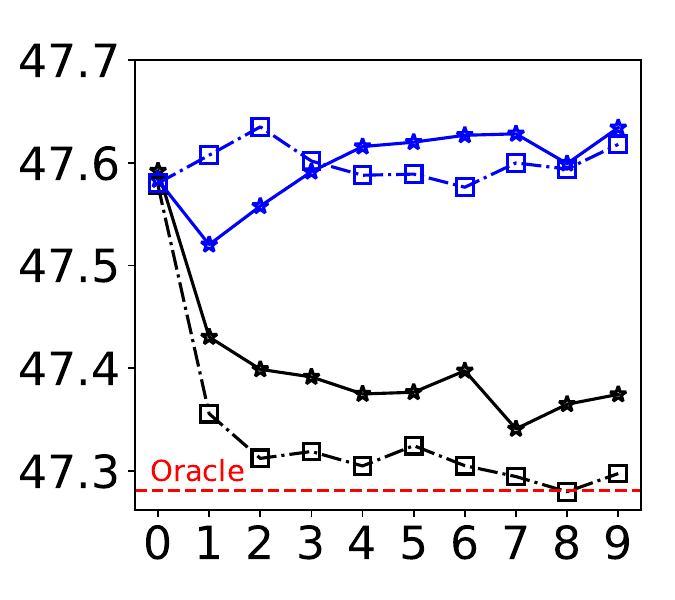} &
        \includegraphics[width=0.19\linewidth]{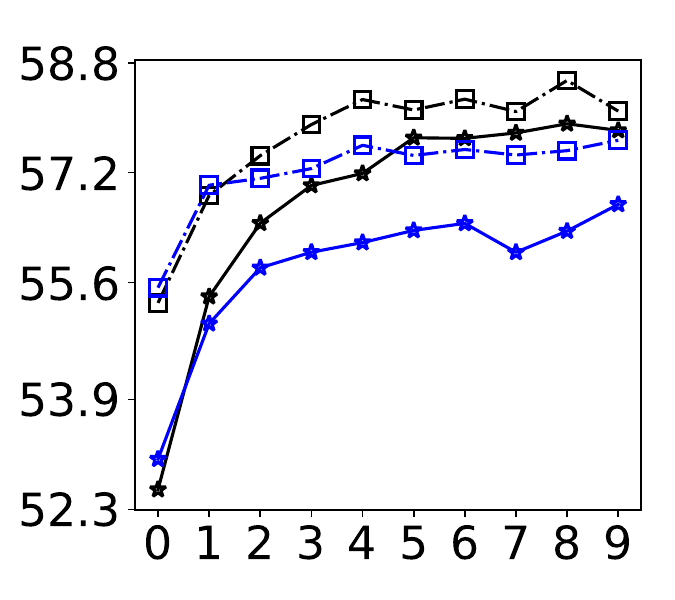} &
        \includegraphics[width=0.19\linewidth]{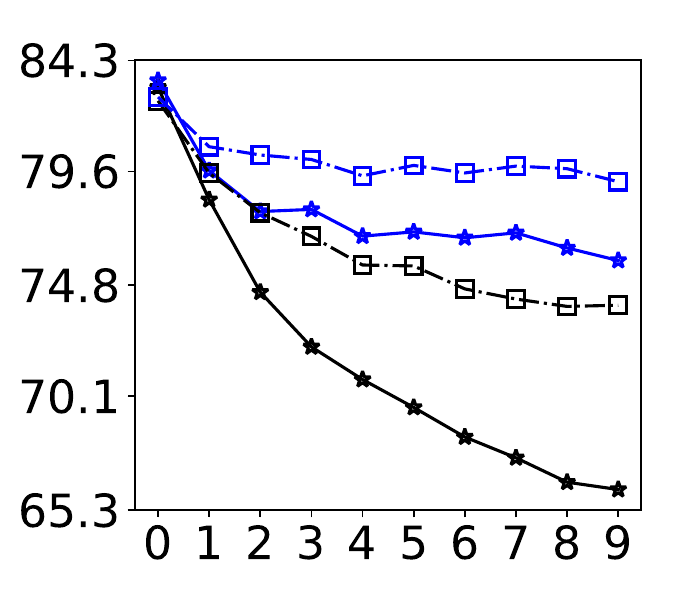} &
        \includegraphics[width=0.19\linewidth]{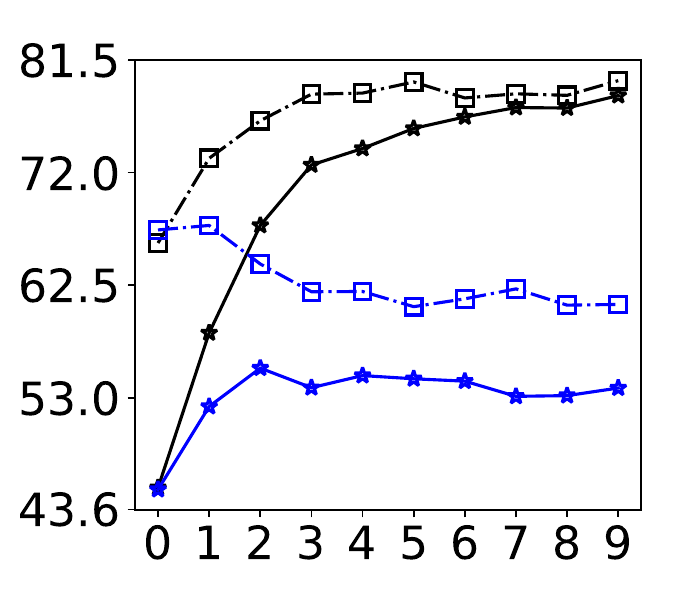} \\
    \end{tabular}
    
    \caption{\texttt{GPT-2} (top) and \texttt{SmolLM2} (bottom) under partially synthetic recursive training $\left(\alpha\!=\!0.5,\beta\!=\!1,\gamma\!=\!0\right)$ for 10 generations. The baseline is equivalent to training on all the data in the pool and the `Oracle' performance represents a perfect AI text detector that filters all synthetic samples. }
    \label{fig:partially_synthetic_a05}
\end{figure*}

\begin{figure*}[!t]
    \centering
    \includegraphics[width=0.85\linewidth]{images/legend.pdf}
    
    \begin{tabular}{@{}c@{\hspace{0.1cm}}c@{}c@{}c@{}c@{}c@{}}
        \raisebox{1.05cm}{\rotatebox{90}{\small \texttt{GPT-2}}} &
        \includegraphics[width=0.19\linewidth]{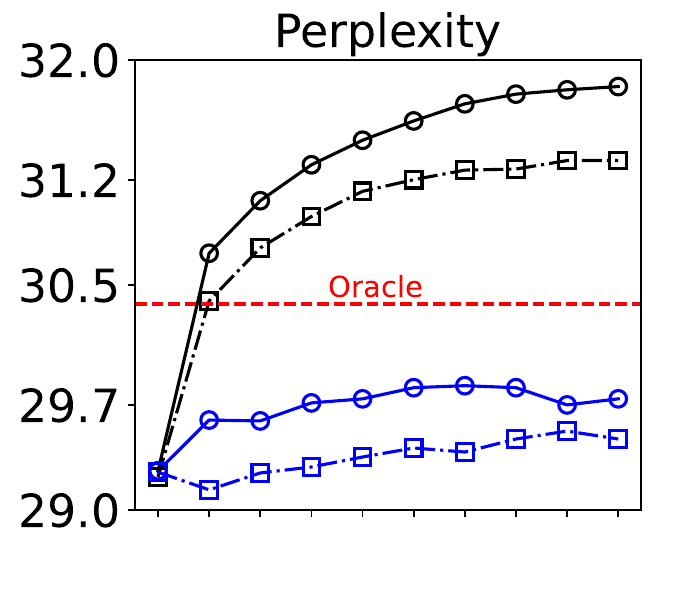} &
        \includegraphics[width=0.19\linewidth]{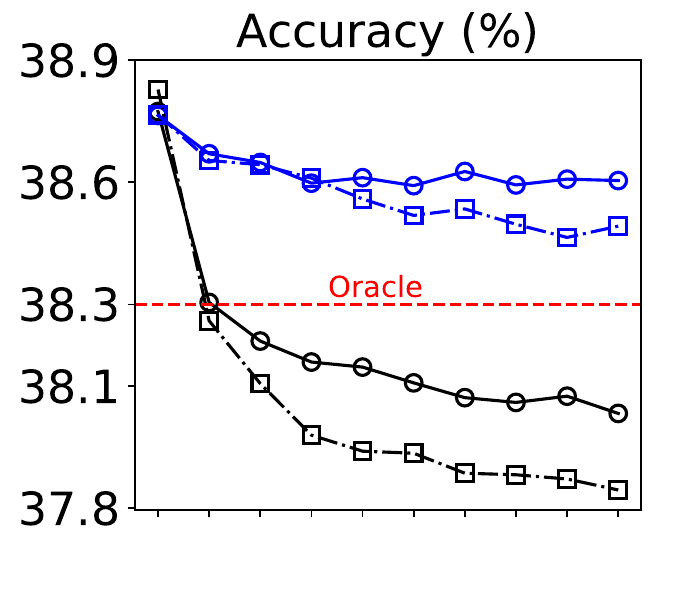} &
        \includegraphics[width=0.19\linewidth]{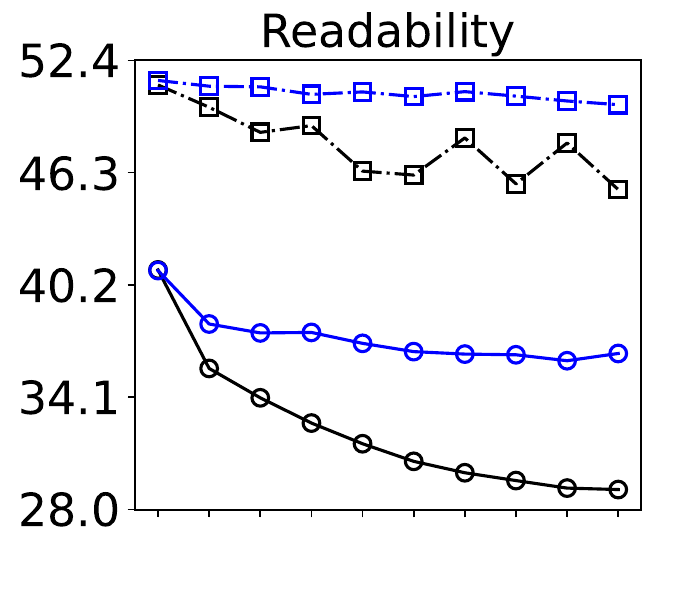} &
        \includegraphics[width=0.19\linewidth]{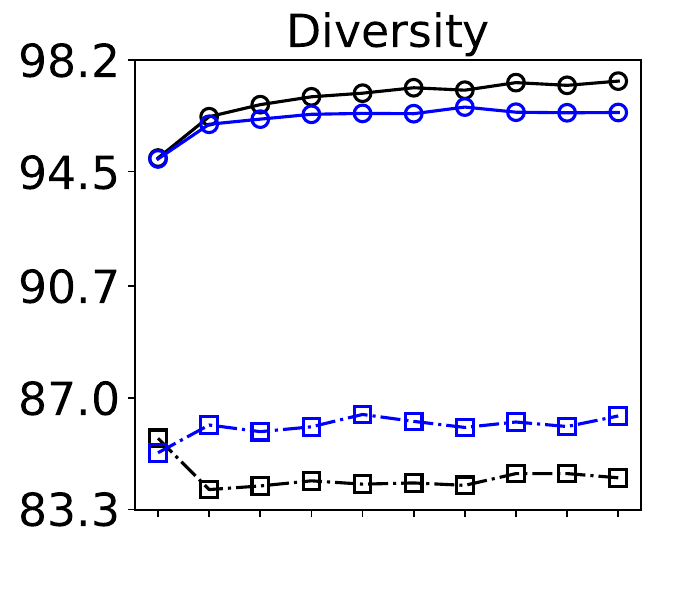} &
        \includegraphics[width=0.19\linewidth]{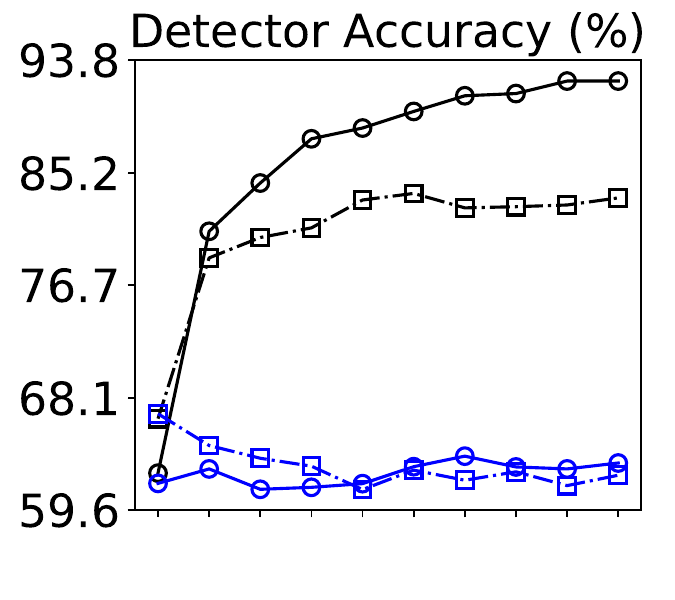} \\[-0.25cm]
        
        \raisebox{0.9cm}{\rotatebox{90}{\small \texttt{SmolLM2}}} &
        \includegraphics[width=0.19\linewidth]{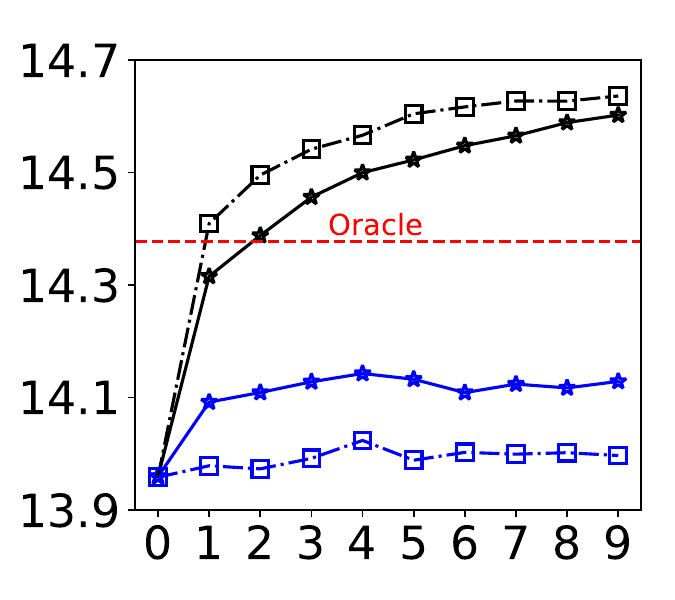} &
        \includegraphics[width=0.19\linewidth]{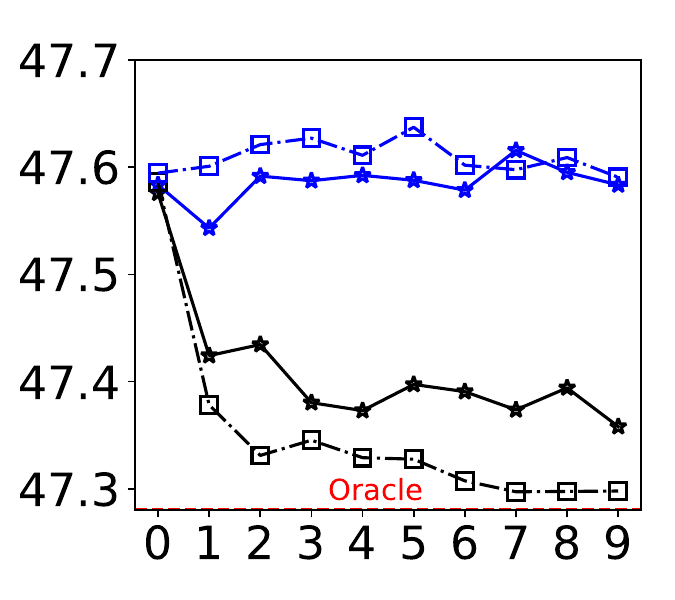} &
        \includegraphics[width=0.19\linewidth]{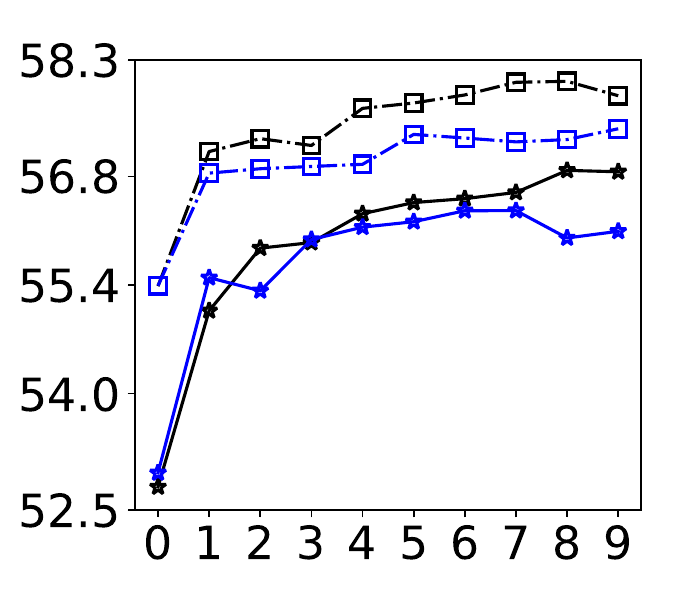} &
        \includegraphics[width=0.19\linewidth]{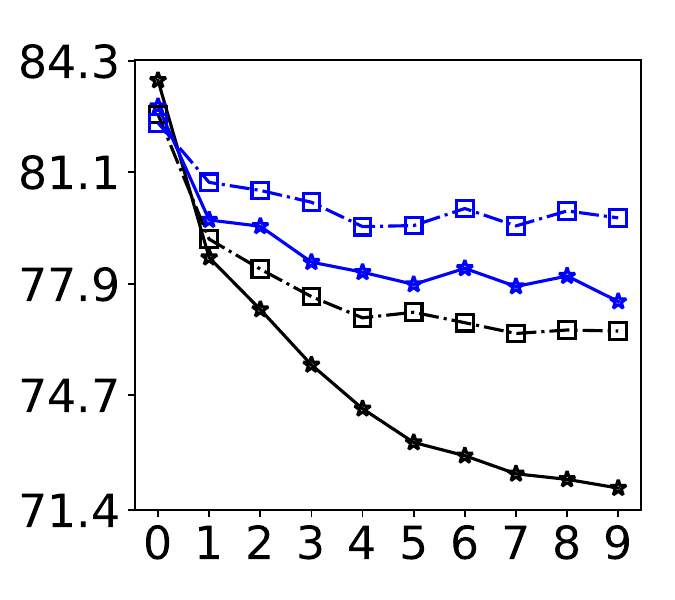} &
        \includegraphics[width=0.19\linewidth]{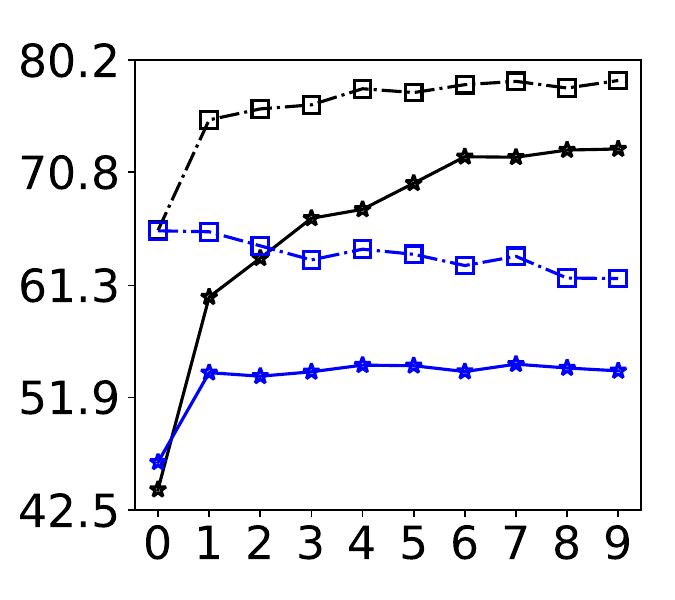} \\
    \end{tabular}
    
    \caption{\texttt{GPT-2} (top) and \texttt{SmolLM2} (bottom) under partially synthetic recursive training with cross-generational data $\left(\alpha\!=\!0.5,\beta\!=\!0.5,\gamma\!=\!0.5\right)$ for 10 generations. The baseline is equivalent to training on all the data in the pool and the `Oracle' performance represents a perfect AI text detector that filters all synthetic samples. }
    \label{fig:partially_synthetic_a05_g05}
\end{figure*}

\begin{figure*}[!t]
    \centering
    \includegraphics[width=0.85\linewidth]{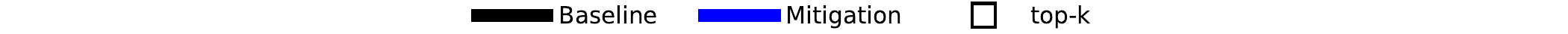}
    \begin{tabular}{@{}c@{\hspace{0.1cm}}c@{}c@{}c@{}c@{}c@{}}
        \raisebox{1.15cm}{\rotatebox{90}{\small \texttt{$1.7$B}}} &
        \includegraphics[width=0.19\linewidth]{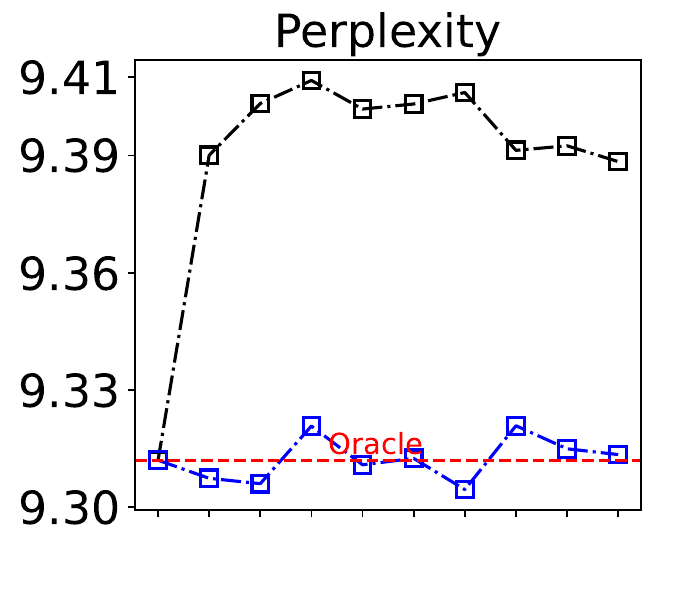} &
        \includegraphics[width=0.19\linewidth]{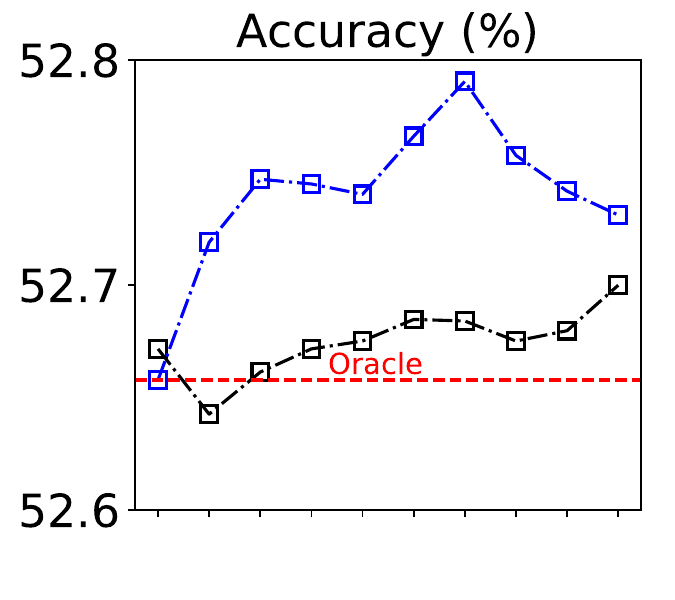} &
        \includegraphics[width=0.19\linewidth]{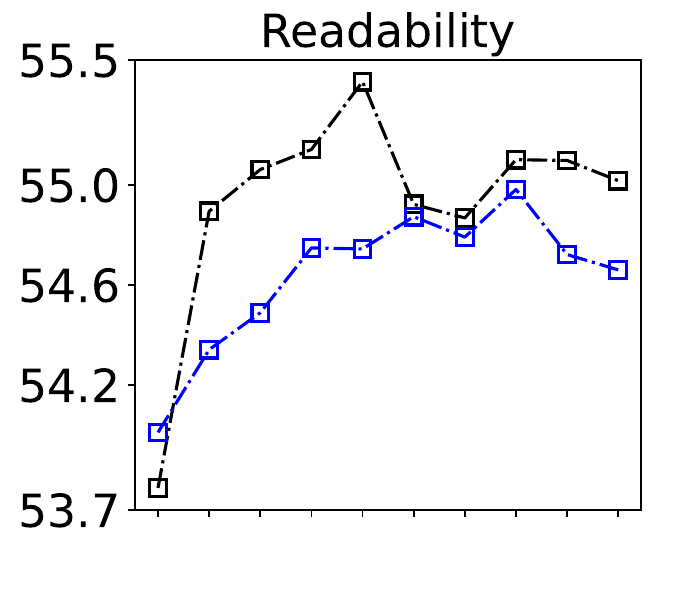} &
        \includegraphics[width=0.19\linewidth]{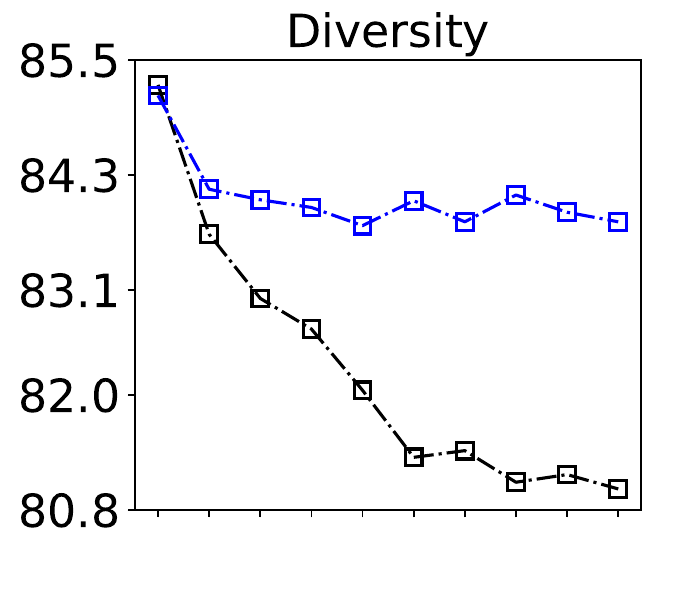} &
        \includegraphics[width=0.19\linewidth]{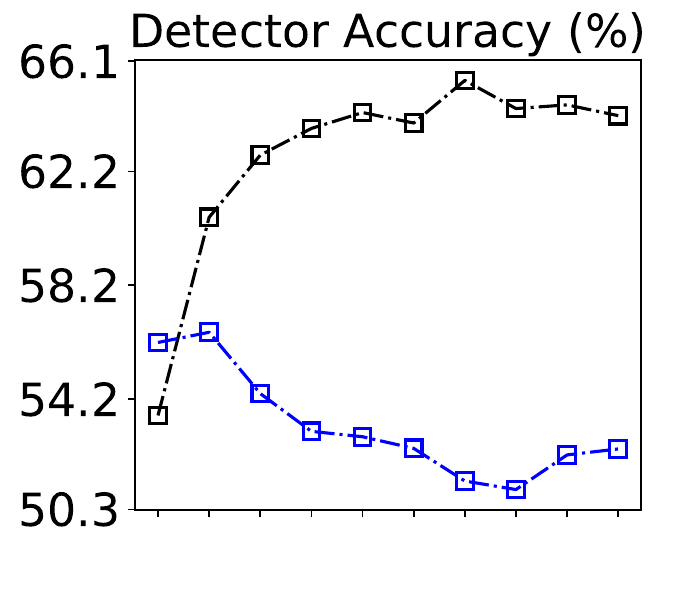} \\
        \raisebox{1.1cm}{\rotatebox{90}{\small \texttt{$360$M}}} &
        \includegraphics[width=0.19\linewidth]{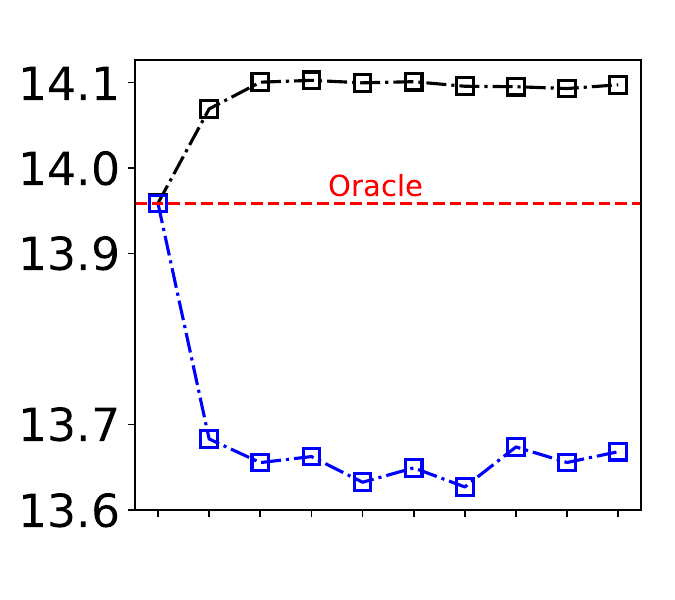} &
        \includegraphics[width=0.19\linewidth]{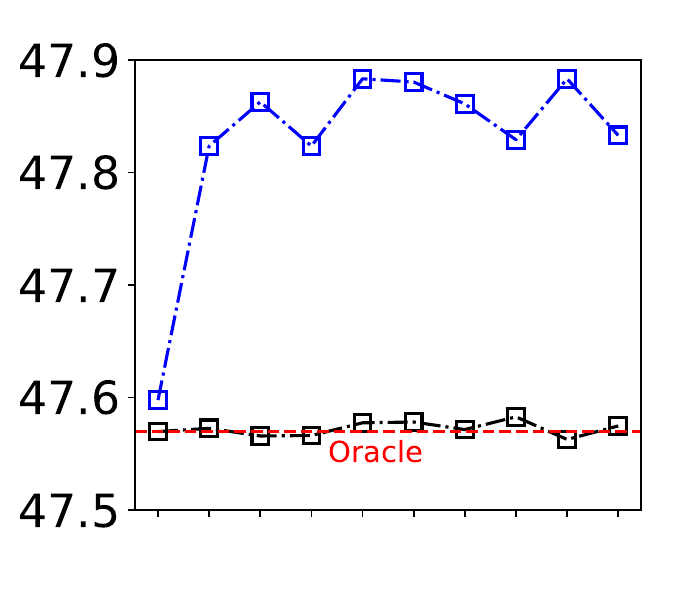} &
        \includegraphics[width=0.19\linewidth]{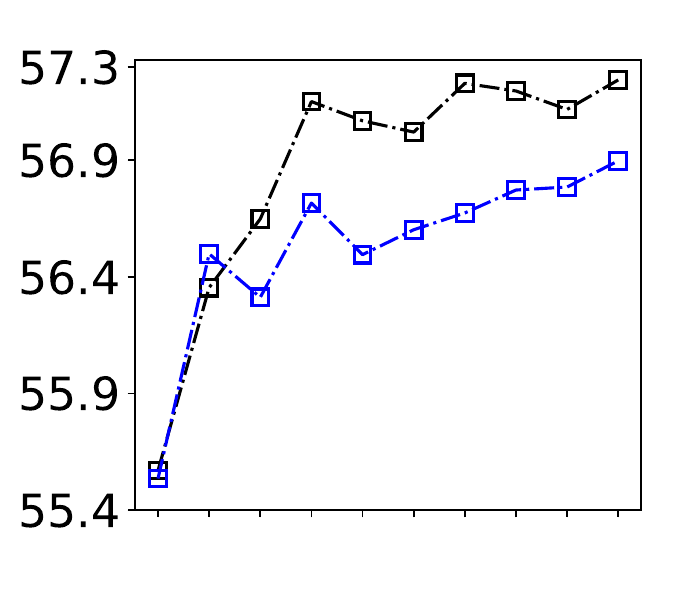} &
        \includegraphics[width=0.19\linewidth]{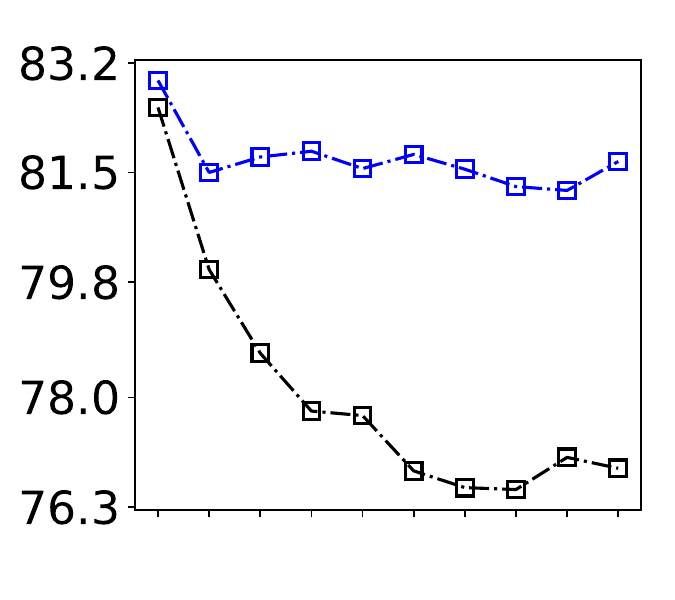} &
        \includegraphics[width=0.19\linewidth]{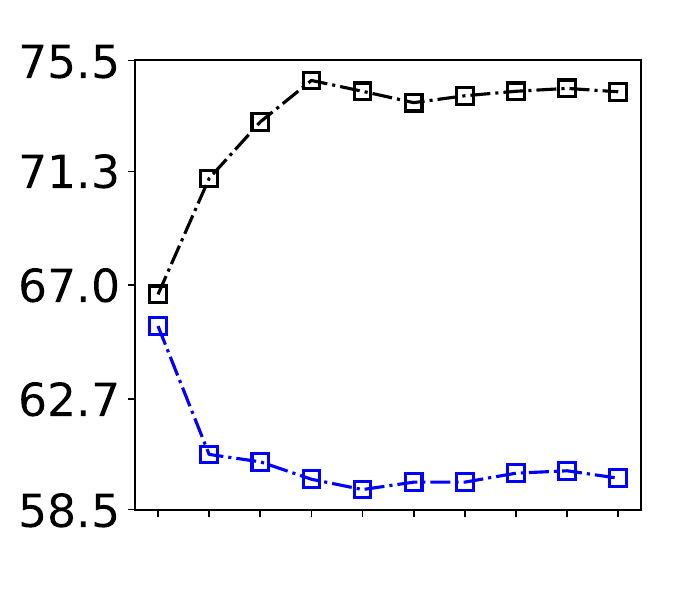} \\
        \raisebox{1.1cm}{\rotatebox{90}{\small \texttt{$135$M}}} &
        \includegraphics[width=0.19\linewidth]{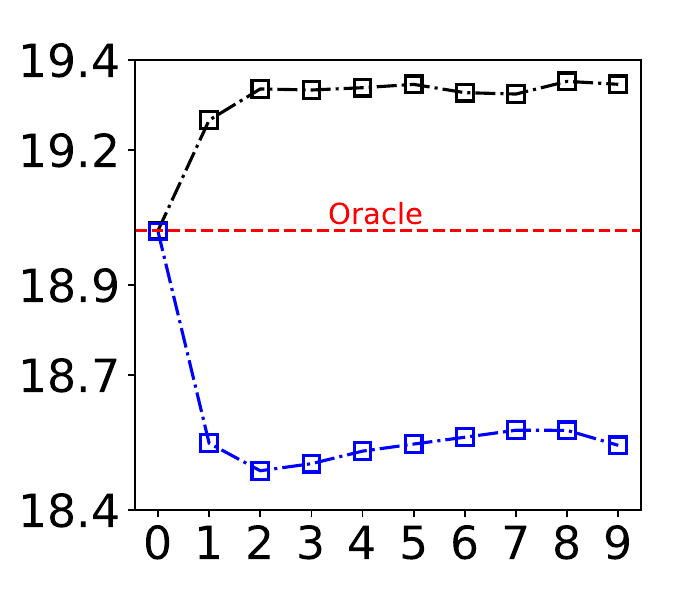} &
        \includegraphics[width=0.19\linewidth]{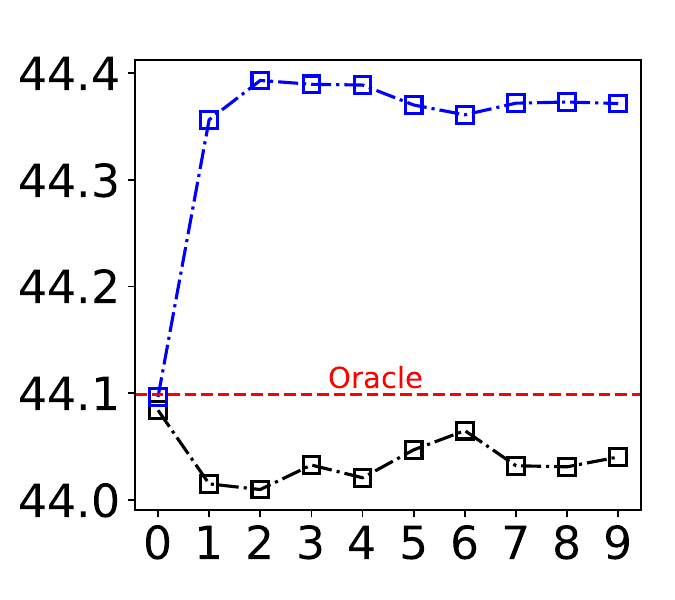} &
        \includegraphics[width=0.19\linewidth]{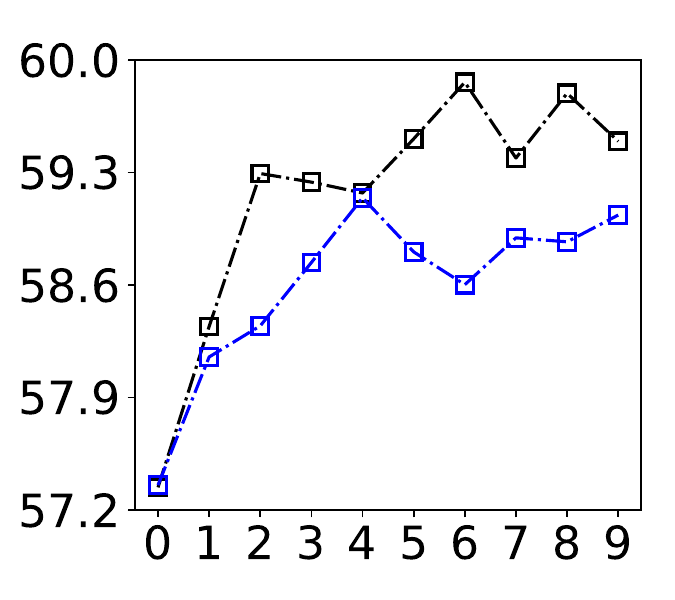} &
        \includegraphics[width=0.19\linewidth]{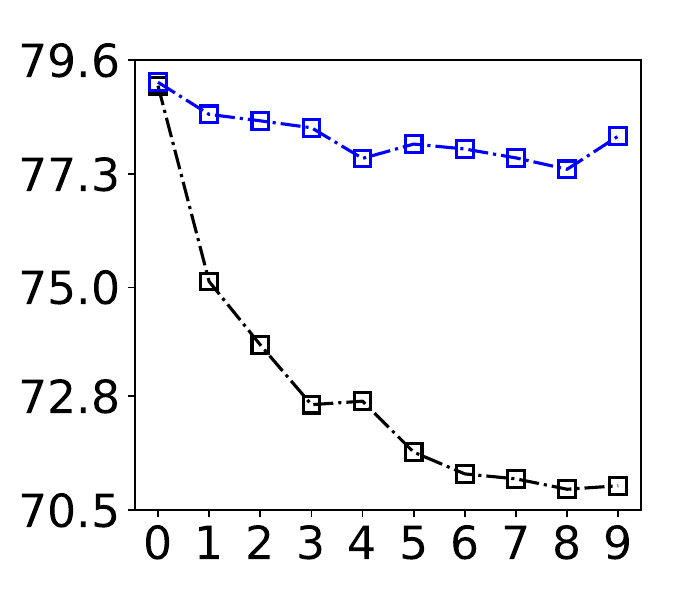} &
        \includegraphics[width=0.19\linewidth]{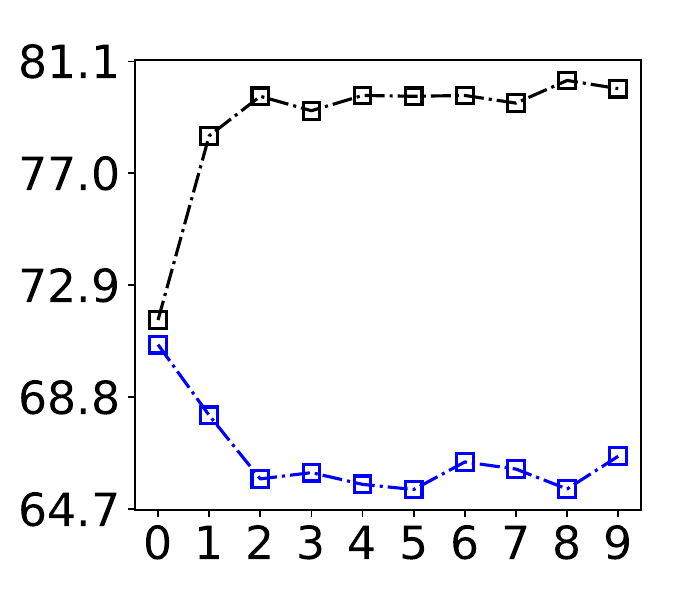} \\
    \end{tabular}
    
    \caption{\texttt{SmolLM2} model size variants ($1.7$B, $360$M, $135$M) under partially synthetic recursive training $\left(\alpha\!=\!1,\beta\!=\!1,\gamma\!=\!0\right)$ for 10 generations. The baseline is equivalent to training on all the data in the pool and the `Oracle' performance represents a perfect AI text detector that filters all synthetic samples. }
    \label{fig:partially_synthetic_model_size}
\end{figure*}

\setlength{\tabcolsep}{1.8pt}
\begin{table*}[t]
    \centering
    \small
    \begin{tabular}{llllcccccccccccc}
        \toprule
        \multirow{2}{*}{\bf Model} & \multirow{2}{*}{\bf Method} & \multirow{2}{*}{\bf Decoding} & \multirow{2}{*}{$\alpha,\beta,\gamma$} & \multicolumn{2}{c}{\bf Perplexity$\downarrow$} & \multicolumn{2}{c}{\bf Accuracy$\uparrow$} & \multicolumn{2}{c}{\bf Diversity$\uparrow$} & \multicolumn{2}{c}{\bf Self-BLEU$\downarrow$} & \multicolumn{2}{c}{\bf MAUVE$\uparrow$} & \multicolumn{2}{c}{\bf Readability$\uparrow$} \\
        & & & & Gen $0$ & Gen $9$ & Gen $0$ & Gen $9$ & Gen $0$ & Gen $9$ & Gen $0$ & Gen $9$ & Gen $0$ & Gen $9$ & Gen $0$ & Gen $9$ \\
        \midrule
        \multirow{12}{*}{\texttt{GPT-2}} 
        & \multirow{6}{*}{baseline}
        & \multirow{3}{*}{top-$k$} 
            & $.5$, $1$, $0$ & $29.23$ & $31.85$ & $38.80$ & $37.56$ & $84.46$ & $82.16$ & $39.10$ & $39.55$ & $93.31$ & $91.10$ & $51.30$ & $46.16$ \\
        & & & $.5$, $.5$, $.5$ & $29.22$ & $31.33$ & $38.83$ & $37.84$ & $85.66$ & $84.35$ & $38.94$ & $40.64$ & $94.07$ & $92.00$ & $51.06$ & $45.38$ \\
        & & & $1$, $1$, $0$ & $29.25$ & $29.92$ & $38.78$ & $38.34$ & $84.66$ & $82.68$ & $39.19$ & $40.20$ & $93.24$ & $92.69$ & $50.99$ & $48.09$ \\
        \cmidrule{3-16}
        & & \multirow{3}{*}{pure sampling} 
            & $.5$, $1$, $0$ & $29.25$ & $32.27$ & $38.78$ & $38.01$ & $94.86$ & $97.77$ & $24.09$ & $14.52$ & $91.08$ & $46.87$ & $40.64$ & $23.69$ \\
        & & & $.5$, $.5$, $.5$ & $29.26$ & $31.82$ & $38.77$ & $38.03$ & $94.95$ & $97.50$ & $24.15$ & $17.02$ & $90.07$ & $76.01$ & $41.02$ & $29.08$ \\
        & & & $1$, $1$, $0$ & $29.25$ & $38.79$ & $38.79$ & $38.63$ & $94.88$ & $96.86$ & $24.04$ & $18.12$ & $91.18$ & $75.64$ & $40.71$ & $31.49$ \\
        \cmidrule{2-16}
        & \multirow{6}{*}{ours}
        & \multirow{3}{*}{top-$k$} 
            & $.5$, $1$, $0$ & $29.25$ & $29.53$ & $38.78$ & $38.45$ & $84.43$ & $85.89$ & $39.43$ & $38.85$ & $94.59$ & $91.52$ & $50.79$ & $50.33$ \\
        & & & $.5$, $.5$, $.5$ & $29.25$ & $29.47$ & $38.77$ & $38.49$ & $85.18$ & $86.40$ & $39.28$ & $38.90$ & $94.75$ & $95.67$ & $51.31$ & $49.99$ \\
        & & & $1$, $1$, $0$ & $29.25$ & $28.59$ & $38.77$ & $38.91$ & $84.81$ & $85.65$ & $38.85$ & $38.76$ & $94.92$ & $93.95$ & $51.28$ & $51.34$ \\
        \cmidrule{3-16}
        & & \multirow{3}{*}{pure sampling}  
            & $.5$, $1$, $0$ & $29.25$ & $29.88$ & $38.79$ & $38.58$ & $94.87$ & $96.50$ & $24.30$ & $19.85$ & $92.92$ & $81.58$ & $40.68$ & $35.59$ \\
        & & & $.5$, $.5$, $.5$ & $29.26$ & $29.74$ & $38.76$ & $38.60$ & $94.92$ & $96.46$ & $24.09$ & $20.66$ & $91.16$ & $88.46$ & $40.98$ & $36.48$ \\
        & & & $1$, $1$, $0$ & $29.24$ & $28.84$ & $38.78$ & $39.00$ & $94.61$ & $96.17$ & $24.01$ & $21.87$ & $91.76$ & $88.06$ & $40.78$ & $38.10$ \\
        \midrule
        \multirow{12}{*}{\shortstack{\texttt{SmolLM2} \\ $350$M}}
        & \multirow{6}{*}{baseline}
        & \multirow{3}{*}{top-$k$} 
            & $.5$, $1$, $0$ & $13.96$ & $14.69$ & $47.58$ & $47.30$ & $82.59$ & $73.95$ & $52.68$ & $54.53$ & $91.58$ & $78.42$ & $55.31$ & $58.10$ \\
        & & & $.5$, $.5$, $.5$ & $13.96$ & $14.64$ & $47.59$ & $47.30$ & $82.76$ & $76.54$ & $52.57$ & $53.97$ & $91.03$ & $86.15$ & $55.38$ & $57.84$ \\
        & & & $1$, $1$, $0$ & $13.96$ & $14.10$ & $47.57$ & $47.57$ & $82.51$ & $76.90$ & $52.47$ & $54.17$ & $91.32$ & $81.11$ & $55.57$ & $57.24$ \\
        \cmidrule{3-16}
        & & \multirow{3}{*}{temperature}  
            & $.5$, $1$, $0$ & $13.96$ & $14.69$ & $47.59$ & $47.37$ & $83.16$ & $66.16$ & $51.21$ & $54.46$ & $83.67$ & $70.15$ & $52.59$ & $57.81$ \\
        & & & $.5$, $.5$, $.5$ & $13.96$ & $14.60$ & $47.58$ & $47.36$ & $83.74$ & $72.03$ & $51.22$ & $54.15$ & $86.94$ & $81.27$ & $52.80$ & $56.86$ \\
        & & & $1$, $1$, $0$ & $13.96$ & $14.06$ & $47.58$ & $47.64$ & $83.31$ & $73.11$ & $51.16$ & $53.96$ & $84.55$ & $76.27$ & $52.74$ & $55.85$ \\
        \cmidrule{2-16}
        & \multirow{6}{*}{ours}
        & \multirow{3}{*}{top-$k$} 
            & $.5$, $1$, $0$ & $13.96$ & $14.02$ & $47.58$ & $47.62$ & $82.75$ & $79.17$ & $52.58$ & $53.89$ & $89.33$ & $90.39$ & $55.53$ & $57.67$ \\
        & & & $.5$, $.5$, $.5$ & $13.96$ & $14.00$ & $47.59$ & $47.59$ & $82.52$ & $79.79$ & $52.64$ & $53.74$ & $88.69$ & $87.92$ & $55.39$ & $57.41$ \\
        & & & $1$, $1$, $0$ & $13.96$ & $13.67$ & $47.60$ & $47.83$ & $82.93$ & $81.67$ & $52.74$ & $53.23$ & $85.09$ & $89.72$ & $55.53$ & $56.90$ \\
        \cmidrule{3-16}
        & & \multirow{3}{*}{temperature}  
            & $.5$, $1$, $0$ & $13.96$ & $14.11$ & $47.58$ & $47.63$ & $83.45$ & $75.84$ & $51.15$ & $53.67$ & $87.80$ & $84.82$ & $53.03$ & $56.74$ \\
        & & & $.5$, $.5$, $.5$ & $13.96$ & $14.13$ & $47.58$ & $47.58$ & $82.99$ & $77.39$ & $51.12$ & $53.08$ & $86.20$ & $81.00$ & $52.98$ & $56.09$ \\
        & & & $1$, $1$, $0$ & $13.96$ & $13.75$ & $47.57$ & $47.83$ & $83.43$ & $79.36$ & $51.21$ & $52.99$ & $88.16$ & $86.94$ & $52.94$ & $55.66$ \\
        \midrule
        \multirow{2}{*}{\shortstack{\texttt{SmolLM2} \\ $135$M}}  & baseline & \multirow{2}{*}{top-$k$} & \multirow{2}{*}{$1$, $1$, $0$} & $19.02$ & $19.35$ & $44.08$ & $44.04$ & $79.08$ & $70.99$ & $53.08$ & $54.87$ & $85.18$ & $72.73$ & $57.34$ & $59.49$ \\
         & ours & & & $19.02$ & $18.54$ & $44.10$ & $44.37$ & $79.15$ & $78.06$ & $52.90$ & $53.69$ & $82.92$ & $82.24$ & $57.35$ & $59.04$ \\
        \midrule
        \multirow{2}{*}{\shortstack{\texttt{SmolLM2} \\ $1.7$B}}  
        & baseline & \multirow{2}{*}{top-$k$} & \multirow{2}{*}{$1$, $1$, $0$} & $9.31$ & $9.39$ & $52.67$ & $52.70$ & $85.24$ & $81.02$ & $53.22$ & $54.28$ & $91.11$ & $85.95$ & $53.79$ & $55.02$ \\
         & ours & & & $9.31$ & $9.31$ & $52.66$ & $52.73$ & $85.13$ & $83.81$ & $53.16$ & $53.68$ & $88.22$ & $93.23$ & $54.01$ & $54.66$ \\
        \bottomrule
    \end{tabular}
    \caption{Test performance (perplexity and accuracy) and data quality at generation $0$ and generation $9$. Results are shown for top-$k$ decoding and pure sampling/temperature for different values of $\alpha$, $\beta$, and $\gamma$ ($\uparrow$ / $\downarrow$: higher / lower is better).}
    \label{tab:mitigation_percentages}
\end{table*}
\end{document}